\pdfoutput=1

\documentclass[11pt]{article}

\usepackage[]{acl}

\usepackage{times}
\usepackage{latexsym}
\usepackage{multirow}
\usepackage{booktabs}
\usepackage{amsmath}
\DeclareMathOperator*{\argmax}{argmax}
\DeclareMathOperator*{\argmin}{argmin}
\DeclareMathOperator*{\sample}{sample}

\usepackage[T1]{fontenc}

\usepackage[utf8]{inputenc}

\usepackage{microtype}

\usepackage{todonotes}

\title{Identifying Weaknesses in Machine Translation Metrics Through Minimum Bayes Risk Decoding: A Case Study for COMET}

\author{Chantal Amrhein$^1$ \and Rico Sennrich$^{1,2}$\\
  $^1$Department of Computational Linguistics, University of Zurich\\
  $^2$School of Informatics, University of Edinburgh \\ \medskip
  \texttt{\{amrhein,sennrich\}@cl.uzh.ch}}

\begin{document}
\maketitle
\begin{abstract}
Neural metrics have achieved impressive correlation with human judgements in the evaluation of machine translation systems, but before we can safely optimise towards such metrics, we should be aware of (and ideally eliminate) biases toward bad translations that receive high scores.
Our experiments show that sample-based Minimum Bayes Risk decoding can be used to explore and quantify such weaknesses. When applying this strategy to COMET for en$\rightarrow$de and de$\rightarrow$en, we find that COMET models are not sensitive enough to discrepancies in numbers and named entities. We further show that these biases are hard to fully remove by simply training on additional synthetic data and release our code and data for facilitating further experiments.\footnote{\url{https://github.com/ZurichNLP/mbr-sensitivity}}
\end{abstract}

\section{Introduction}

Recently, neural machine translation evaluation metrics have reached better correlation scores with human evaluators than surface-level metrics like BLEU \citep{papineni-etal-2002-bleu}. In particular, COMET \citep{rei-etal-2020-comet} has shown significant potential as a leading evaluation metric both in shared tasks \citep{mathur-etal-2020-results, freitag-etal-2021-results} and other studies on machine translation evaluation metrics \citep{kocmi-etal-2021-ship}. The main benefits of such neural metrics are that they do not rely on surface-level similarity to a reference translation and that some of them operate in a multilingual representation space. This also allows for comparing translations to the source sentence.

A recent evaluation as part of the WMT 2021 metrics shared task \citep{freitag-etal-2021-results} suggests that neural metrics are also less susceptible to many weaknesses of earlier non-neural metrics, e.g.\ an antonym in the translation hurting the BLEU score exactly the same amount as a synonym. However, it is still unclear whether or not these metrics also introduce new biases that are harder to detect since they are essentially ``black box'' metrics that do not explain why a certain score is attributed to a translation.
Failing to identify these biases in neural metrics could lead the community to optimise towards metric ``blind spots'', either directly through reward-based training methods such as Minimum Risk Training \cite{shen-etal-2016-minimum}, or more slowly by basing modelling choices on metric scores.
It is therefore worthwhile to find new means to uncover weaknesses of neural machine translation metrics. 

In this paper, we show that sampling-based Minimum Bayes Risk (MBR) decoding - where a pool of samples are compared against each other using a machine translation evaluation metric as a utility function - can render blind spots of these metrics more observable. When applying COMET as the utility function, we find many examples where a translation hypothesis is chosen that contains different numbers or named entities than the source and reference (see examples in Table \ref{tab:examples}). Through a targeted sensitivity analysis, we identify that these are indeed weaknesses of COMET and we show that it can be hard to remove them from the model.

Our contributions are the following:

\begin{itemize}
    \item We propose to use sample-based MBR decoding to explore and measure weaknesses of neural machine translation evaluation metrics.
    \item We find that COMET is not sensitive enough to number differences and mistranslations of named entities when translating from de$\leftrightarrow$en.
    \item We show that simply retraining COMET on synthetic data is not enough to fully eliminate these blind spots.
\end{itemize}

\begin{table*}[ht]
    \centering
    
    \begin{tabular}{lc}
        \small \texttt{src} & \small Schon drei Jahre nach der Gründung verließ Green die Band \textbf{1970}. \\
        \small \texttt{ref} & \small Green left the band three years after it was formed, in \textbf{1970}. \\
        \small \texttt{MBR\textsubscript{chrF++}} &  \small Already three years after the foundation, Green left the band in \textbf{1970}.  \\ 
        \small \texttt{MBR\textsubscript{COMET}} &  \small Three years after the creation, Green left the band in \colorbox[HTML]{F9CBD0}{\textbf{1980}}. \\ \\ \addlinespace

        \small \texttt{src} & \small [...] \textbf{Mahmoud} Guemama's Death - Algeria Loses a Patriot [...], Says President \textbf{Tebboune}.\\
        \small \texttt{ref} & \small  [...] \textbf{Mahmoud} Guemamas Tod - Algerien verliert einen Patrioten [...], sagt Präsident \textbf{Tebboune}.\\
        \small \texttt{MBR\textsubscript{chrF++}} &  \small  [...] \textbf{Mahmoud} Guemamas Tod - Algerien verliert einen Patriot [...], sagt Präsident \textbf{Tebboune}.\\
        \small \texttt{MBR\textsubscript{COMET}} &  \small [...] \colorbox[HTML]{F9CBD0}{\textbf{Mahmud}} Guemamas Tod - Algerien verliert einen Patriot [...], sagt Präsident \colorbox[HTML]{F9CBD0}{\textbf{Tebboene}}. \\ 
         
    \end{tabular}
    \caption{Examples of MBR decoding outputs with chrF++ and COMET as utility metrics. The outputs chosen with COMET indicate less sensitivity towards discrepancies in numbers and named entities. }
    \label{tab:examples}
\end{table*}

\section{Related Works}

How to best evaluate machine translation models has been a long-standing question in the research community. Ideally, we could employ humans to judge the quality of different models but this is time-consuming, costly and requires trained professionals. Various automatic machine translation metrics have been proposed over the years that typically compare a machine translation output to a reference sentence according to surface-level similarity \citep{papineni-etal-2002-bleu, popovic-2015-chrf} or on a shallow semantic level \citep{banerjee-lavie-2005-meteor}. 

With the rise of contextual embeddings and large multilingual Transformer language models, metrics that map translations and references into the same latent space and compare the cosine similarity between them \citep{lo-2020-extended} or use them as inputs to predict a score \citep{sellam-etal-2020-bleurt, rei-etal-2020-comet} have become popular. Such neural metrics have been shown to agree more with human evaluation than previously popular metrics such as BLEU \citep{papineni-etal-2002-bleu} or chrF \citep{popovic-2015-chrf}.

However, these neural metrics can also introduce new biases that we are not yet aware of \citep{hanna-bojar-2021-fine}. In this paper, we aim to find a way to identify such weaknesses via Minimum Bayes Risk (MBR) decoding. While MBR decoding was a frequently used decoding strategy in the days of statistical machine translation \citep{GOEL2000115, kumar-byrne-2004-minimum, tromble-etal-2008-lattice}, it has only recently gained traction in the context of neural machine translation. \citet{eikema-aziz-2020-map} argue that MBR decoding using samples as hypotheses results in an unbiased candidate pool in contrast to beam search outputs which maximise the probability under the model. Indeed, if the machine translation model generating the samples is strong enough, humans prefer MBR-decoded hypotheses selected with BLEURT \citep{sellam-etal-2020-bleurt} as the utility function over beam search outputs \citep{freitag-etal-2022-high}. 

\citet{muller-sennrich-2021-understanding} further show that MBR outputs can inherit biases from the utility function, for example, the length bias \citep{nakov-etal-2012-optimizing} when BLEU is used as the utility function. Consequently, it stands to reason that MBR decoding can also be used to uncover new biases of metrics that are used as utility functions, as we will show in this work.

\section{Minimum Bayes Risk Decoding}
\label{sec:mbr}
Traditionally, maximum a posteriori (MAP) decoding is used in the context of neural machine translation. The goal is to find the translation hypothesis $h_i$ among all possible hypotheses $H$ that is most probable under the translation model given the source sentence $x$ and the model parameters $\theta$:

\begin{equation}
    y^* = \argmax_{h_i \in H} p_{model}(h_i|x,\theta)
\end{equation}

In practice, it is not feasible to consider every possible hypothesis. Beam search offers a popular and effective approximation.

In contrast, MBR decoding aims to find a translation that minimises the expected cost (risk) of choosing a candidate translation $h_i$, assuming that we have some loss function $L$ to compare the candidate to a true translation $h_j$, and access to the true probability distribution $P$:

\begin{equation}
    y^* = \argmin_{h_i \in H} \sum_{h_j \in H} P(h_j|x) L(h_i, h_j)
\end{equation}

Since we do not have access to the true probability distribution $P$, and cannot exhaustively sum over all possible translations $H$, we have to make several approximations.
First, we select a subset of all possible hypotheses $H$ as candidate translations $C$ to make the computation tractable. \citet{eikema-aziz-2020-map} suggest drawing ancestral samples from the translation model as a set of unbiased candidates, and we follow this sampling-based MBR approach. Ancestral samples $s$ are created by sampling the next token $w$ from the translation model according to the probability distribution over the vocabulary $V$ at each time step $t$:

\begin{equation}
    s_t = s_{t-1} + \sample_{w_i \in V} (p_{model}(w_i|x,s_{t-1},\theta))
\end{equation}

The probability distribution is conditioned on the source sentence $x$ and the previously produced output tokens $s_{t-1}$. For each ancestral sample $s$, this sampling continues until the end-of-sentence symbol is sampled as the next token $w$.

Second, we need to create an additional set of ``support hypotheses'' $S$ that serve as an approximation to the unknown true translation. The set of candidates $C$ and the set of support hypotheses $S$ can be created separately but in this work, we follow \citet{eikema-aziz-2020-map} and let our translation model produce a set of 100 ancestral samples that are used both as candidates and support ($C = S$).

Third, we need to define a loss function $L$. In practice, we often substitute the loss function for a similarity function where higher values are better. Such a ``utility function'' $u$ is then used to search for the translation $h_i$ that maximises the expected utility or – to paraphrase – is most similar to all hypotheses in the support set $S$:

\begin{equation}
    y^* = \argmax_{h_i \in C} \frac{1}{|S|} \sum_{h_j \in S} u(h_i, h_j)
\end{equation}

Any automatic machine translation evaluation metric can be used as the utility function $u$. \citet{eikema2021samplingbased} find that BEER \citep{stanojevic-simaan-2014-fitting} works best among a range of non-neural metrics. More recently, \citet{freitag-etal-2022-high} compare several metrics as utility functions in a human evaluation of MBR-decoded outputs where the neural metric BLEURT \citep{sellam-etal-2020-bleurt} clearly outperforms non-neural metrics. In this paper, we explore the use of another neural evaluation metric as the utility function, namely COMET. Since the reference-based COMET model takes the source, a translation hypothesis and a reference (approximated in MBR decoding with another hypothesis) as input, our formulation of MBR decoding now takes into account the source sentence $x$:

\begin{equation}
    y^* = \argmax_{h_i \in C} \frac{1}{|S|} \sum_{h_j \in S} u(x, h_i, h_j)
\end{equation}

For an efficiency-related discussion of our implementation, please refer to Section \ref{implementation}.

\section{Experiment Setup}
\label{sec:setup}

\subsection{Translation Model}
To be able to generate samples, we train two Transformer Base machine translation models \citep{NIPS2017_7181} using the \texttt{nematus}\footnote{\url{github.com/EdinburghNLP/nematus}} \citep{sennrich-etal-2017-nematus} framework, one from de$\rightarrow$en and one from en$\rightarrow$de. We follow \citet{eikema2021samplingbased} and use all available parallel data from the WMT 2018 news shared task \citep{bojar-etal-2018-findings} except for Paracrawl as training data. This amounts to 5.9 million sentence pairs. After deduplication, we have approximately 5.6 million training examples.

Both models are trained for 250k updates and we choose the best checkpoint based on the BLEU score as evaluated on \texttt{newstest2017} using SacreBLEU \citep{post-2018-call}. We compute a joint subword vocabulary of size 32k with byte pair encoding \citep{sennrich-etal-2016-neural} using the  SentencePiece implementation \citep{kudo-richardson-2018-sentencepiece}. During training and decoding, the maximum sequence length is set to 200 tokens.

Our models are built with 6 encoder layers, 6 decoder layers, 8 attention heads with an embedding and hidden state dimension of 512 and a feed-forward network dimension of 2048. For regularisation, we use a dropout rate of 0.1 for BPE-dropout \citep{provilkov-etal-2020-bpe} during training, for the embeddings, for the residual connections, in the feed-forward sub-layers and for the attention weights. We train with tied encoder and decoder input embeddings as well as tied decoder input and output embeddings \citep{press-wolf-2017-using} and apply exponential smoothing of model parameters (decay $10^{-4}$) \cite{junczys-dowmunt-etal-2018-marian}. Following previous work on MBR decoding \citep{eikema-aziz-2020-map}, we train without label smoothing.

For optimisation, we use Adam \citep{DBLP:journals/corr/KingmaB14} with standard hyperparameters and a learning rate of  $10^{-4}$. We follow the Transformer learning schedule described in \citep{NIPS2017_7181} with a linear warm-up over 4,000 steps. Our token batch size is set to 16,348 and we train on 4 NVIDIA Tesla V100 GPUs.

\subsection{COMET Models}
We experiment with two COMET models that were trained towards two different regression objectives:
\begin{itemize}
    \item \texttt{wmt20-comet-da} \citep{rei-etal-2020-unbabels}, developed for the WMT 2020 metrics shared task \citep{mathur-etal-2020-results} and trained to predict Direct Assessment (DA) \citep{graham_baldwin_moffat_zobel_2017} scores.
    \item \texttt{wmt21-comet-mqm} \citep{rei-etal-2021-references}, developed for the WMT 2021 metrics shared task \citep{freitag-etal-2021-results} and trained to predict MQM scores \citep{freitag-etal-2021-experts} based on the Multidimensional Quality Metrics (MQM) methodology \citep{uszkoreit2013multidimensional}.
\end{itemize}

\subsection{MBR Decoding Implementations}
\label{implementation}
For non-neural metrics, we use the MBR decoding implementation\footnote{\url{https://github.com/Roxot/mbr-nmt}} provided by \citet{eikema2021samplingbased}. We use only unique samples such that no hypothesis is assigned a higher average MBR score simply because it perfectly matches one or multiple hypotheses in the support.\footnote{Using all samples does not affect our results.} 
In our experiments, we use chrF++ \citep{popovic-2017-chrf} and BLEU as non-neural metrics. For BLEU, the implementation internally uses SacreBLEU \citep{post-2018-call}\footnote{Using floor smoothing with a smoothing value of 0.1.}.

For our experiments with COMET, we adapt the official COMET implementation\footnote{\url{https://github.com/Unbabel/COMET}} and implement an option for MBR decoding. Since COMET first creates a pooled sentence representation of the source and each of the two hypotheses before constructing a single vector from these representations and passing it through a regression layer, it is crucial that the implementation does not naively call COMET on every hypothesis pair. Instead, we encode the source sentence and hypotheses \textbf{only once} with XLM-R \citep{conneau-etal-2020-unsupervised} and then score all combinations of hypothesis pairs in parallel. 

\subsection{Evaluation Data}
We decide to use the test sets from the WMT 2021 news shared task \citep{akhbardeh-etal-2021-findings} as our evaluation data. This dataset brings two major benefits to our analysis:
\begin{itemize}
    \item In the de$\leftrightarrow$en directions, it provides at least two references for every source sentence. This allows us to compare how much MBR scores differ between two equivalent human translation alternatives as a reference point. 
    \item This dataset was not part of the training data of the \texttt{wmt20-comet-da} and \texttt{wmt21-comet-mqm} COMET models which avoids the risk that the models have seen scores for similarly erroneous translations of these source sentences before. 
\end{itemize}

There are 1000 sentence triplets (source, two human translations) for de$\rightarrow$en where we use translation A as our reference and translation B as an alternative translation and 1002 sentence triplets for en$\rightarrow$de where we use translation C as our reference and translation D as an alternative translation.

\begin{table*}[t]
    \centering
    \small
    \begin{tabular}{cccccccccc}
        &  \multicolumn{4}{c}{Numbers} &  \multicolumn{4}{c}{Named Entities}\\ 
        \cmidrule(lr){2-5} \cmidrule(lr){6-9} \addlinespace
         &  \multicolumn{2}{c}{\texttt{de-en}} &  \multicolumn{2}{c}{\texttt{en-de}} &  \multicolumn{2}{c}{\texttt{de-en}}  & \multicolumn{2}{c}{\texttt{en-de}} \\ 
        \cmidrule(lr){2-3}  \cmidrule(lr){4-5}  \cmidrule(lr){6-7}   \cmidrule(lr){8-9} \addlinespace
        reference &  93.24  &   & 93.46 & &  n/a & & n/a &\\\addlinespace 
        alternative &  94.83 & +\phantom{0}1.59 &  95.66 & +\phantom{0}2.20 & 73.73 &  & 77.66 &\\\addlinespace 
        beam search & 95.91 & +\phantom{0}2.67 & 95.73 & +\phantom{0}2.27 & 71.55 & -\phantom{0}2.18 & 70.03 & -\phantom{0}7.63\\\addlinespace 
        MBR chrF++ &  91.22 & -\phantom{0}2.02 &  93.43 & -\phantom{0}0.03 & 67.59 & -\phantom{0}6.14 & 62.44 & -15.22\\\addlinespace 
        MBR bleu &  93.88 &  +\phantom{0}0.64 &  91.37 & -\phantom{0}2.09 & 65.14 & -\phantom{0}8.59 & 62.50 & -15.16\\\addlinespace 
        MBR \small \texttt{wmt20-comet-da} &  90.34 & \textbf{-\phantom{0}2.90} &  89.14 & \textbf{-\phantom{0}4.32} & 65.33 & \textbf{-\phantom{0}8.40} & 54.17 & \textbf{-23.49}\\\addlinespace 
        MBR \small \texttt{wmt21-comet-mqm} &  82.35 & \textbf{-10.89} &  77.10 & \textbf{-16.36} & 58.15 & \textbf{-15.58} & 53.31 & \textbf{-24.35}\\\addlinespace \hline \addlinespace
        MBR \small \texttt{retrain-comet-da} &  \colorbox[HTML]{B2EAB1}{92.65} & \colorbox[HTML]{B2EAB1}{-\phantom{0}0.59} &  \colorbox[HTML]{B2EAB1}{90.17} & \colorbox[HTML]{B2EAB1}{-\phantom{0}3.29} & \colorbox[HTML]{B2EAB1}{66.48} & \colorbox[HTML]{B2EAB1}{-\phantom{0}7.25} & \colorbox[HTML]{B2EAB1}{60.48} & \colorbox[HTML]{B2EAB1}{-17.18}\\
    \end{tabular}
    \caption{Results of the automatic evaluation. F1-scores (\%) for number and named entity matches and F1-score changes compared to the reference for numbers and alternative translation for named entities. F1-scores that increased after retraining COMET are marked in green.}
    \label{tab:automatic}
\end{table*}

\section{Exploration of MBR-Decoded Outputs}
\label{sec:exploration}
We employ sampling-based MBR decoding as a strategy to identify weaknesses in evaluation metrics that are used as utility functions. We believe that – in addition to general errors – we may also find other errors that can stem from two sources:

First, since samples are often of lower quality than hypotheses produced with beam search, neural metrics may behave unexpectedly when faced with errors that occur less frequently in beam search based machine translation outputs on which they were trained.
Second, in MBR decoding, we compare a candidate translation hypothesis to a pseudo-reference (another hypothesis) instead of an actual reference. This is also something neural metrics were neither trained on nor designed to do.

We are most interested in general errors and errors of the first type since the second type is only relevant for MBR decoding itself. Therefore, we conduct additional experiments in Section \ref{sec:sensitivity} to distinguish between these two sources for the errors we identify below. Note that errors of the second type may become more important to investigate as MBR decoding becomes more prevalent or if we evaluate against multiple translation hypotheses instead of references \citep{fomicheva-etal-2020-multi}.

In our experiments, we first manually compare MBR-decoded outputs that were chosen with two different evaluation metrics as the utility function: chrF++ and COMET. For COMET, we notice several cases where the chosen hypothesis contains numbers and named entities that do not match with the source and the reference, even though the majority of samples in the support set contain the correct numbers and named entities. Two examples are shown in Table \ref{tab:examples}.

To test if these findings apply at scale, we run an automatic evaluation. For numbers, we use regular expressions to identify numbers in the MBR-decoded outputs. We measure the overlap between numbers in the source and the translation with the F1-score.
We decide to compare to the source to be able to compute the overlaps for the reference and the alternative human translation as well. The results can be seen in the left part of Table \ref{tab:automatic}. For named entities, we use \texttt{spaCy}\footnote{English: en\_core\_web\_lg, German: de\_core\_news\_lg}  \citep{spacy} to identify entities of type ``person''. Here, we compute the F1-scores to measure the overlap to the reference rather than to the source (as done for numbers) since the named entity recognition (NER) models are different for English and German. The results are shown in Table \ref{tab:automatic} on the right. 

These simple automatic ``gold'' annotations produce false positives\footnote{For example, translating ``3 pm'' in the source to ``15:00'' is a valid translation, but would be counted as a mistake with the automatic number matching. Similarly, numbers translated as numerals are counted as errors, e.g. ``15'' and ``fifteen''.}, which explains why neither the reference nor the alternative reference (for named entities) achieves an F1-score of 100\%. However, this approximate method is sufficient to expose the large gap between the reference translation, the beam search output, and the output with MBR decoding with surface-level metrics and with COMET. We perform a manual error analysis of all numbers that our evaluation script identifies as errors for the MBR decoded outputs. The false positive rate is similar for all three utility functions: Around 3\% of all numbers that occur either in the source or the translation are mistakenly identified as number mismatches. In contrast, the percentage of genuine errors increases: MBR bleu has a true negative rate of 4.4\%, MBR chrF++ of 4.6\%, MBR \texttt{wmt20-comet-da} of 7.2\% and MBR \texttt{wmt21-comet-mqm} of 16.6\%  (computed jointly over de$\leftrightarrow$en). Thus, the wide gap caused with COMET as the utility function is due to genuine number mismatches, not paraphrasing.

Consequently, these results indicate that MBR decoding with the COMET metrics chooses more erroneous translations with respect to these criteria than with the two non-neural metrics or compared to beam search decoding. Interestingly, the \texttt{wmt21-comet-mqm} model performs considerably worse than the  \texttt{wmt20-comet-da} model in this analysis. Oracle experiments where we choose the sample closest to the two references according to different metrics (see Appendix \ref{app:oracle}) show smaller F1-score differences between both COMET models and the non-neural metrics but they still perform worse, particularly compared to chrF++.

It is worth noting that the beam search output has the highest F1-score of all tested decoding strategies. This suggests that mistranslations of numbers and named entities do not occur as frequently in beam search outputs and COMET's insensitivity to numbers and named entities could therefore be less harmful when evaluating beam search outputs. However, \citet{wang-etal-2021-easy} recently showed that state-of-the-art research models and commercial NMT systems still struggle with numerical translations even when decoding with beam search. Such mistranslations may also occur more frequently in out-of-domain and low-resource settings and therefore, we argue that this insensitivity of COMET is not only harmful for sampling-based MBR decoding but also when evaluating beam search output. 

This automatic evaluation has strengthened the findings in our manual exploration that wrong number and named entity translations are recurring problems. To better quantify how sensitive COMET models are toward these error types, we propose to perform an MBR-based sensitivity analysis in the next section.

\begin{table*}[ht]
    \centering
    \small
    \begin{tabular}{lrccccccrcc}
         &&  \multicolumn{3}{c}{Samples as Support} & \multicolumn{3}{c}{References as Support} & & \multicolumn{2}{c}{Controls}\\ \cmidrule(lr){3-5} \cmidrule(lr){6-8} \cmidrule(lr){10-11} 
         && Numbers & NEs & Nouns &  Numbers & NEs & Nouns & & Samples & Ref.\\           \cmidrule(lr){3-3}  \cmidrule(lr){4-4}   \cmidrule(lr){5-5} \cmidrule(lr){6-6}  \cmidrule(lr){7-7}   \cmidrule(lr){8-8}   \cmidrule(lr){10-10} \cmidrule(lr){11-11} 
          \\
         \multirow{4}{*}{\textbf{de-en}} & add & -0.047 & -0.054 & \colorbox[HTML]{F9CBD0}{-0.255} & -0.086 & -0.101 & \colorbox[HTML]{F9CBD0}{-0.385} & altern. & \phantom{-}0.022 & \\
         &del & -0.048 & -0.044 & \colorbox[HTML]{F9CBD0}{-0.214}  & -0.085 & -0.079 & \colorbox[HTML]{F9CBD0}{-0.314} & copy & -0.593 & -0.472\\
         &sub  & -0.024 & -0.056 & \colorbox[HTML]{F9CBD0}{-0.270} & -0.041 & -0.119 & \colorbox[HTML]{F9CBD0}{-0.410}& hallucin. & -1.277  & -1.907\\
         &whole & -0.064 & -0.122 & \colorbox[HTML]{F9CBD0}{-0.320} & -0.111 & -0.212 & \colorbox[HTML]{F9CBD0}{-0.496} & \\\addlinespace
         \multirow{4}{*}{\textbf{en-de}} & add & -0.024 & -0.053 &  \colorbox[HTML]{F9CBD0}{-0.160} & -0.057 & -0.108 & \colorbox[HTML]{F9CBD0}{-0.257} & altern. & -0.014 \\
         &del  & -0.037 & -0.044 & \colorbox[HTML]{F9CBD0}{-0.113}  & -0.063 & -0.078 & \colorbox[HTML]{F9CBD0}{-0.215} & copy & -1.449 & -1.350\\
         &sub  & -0.011 & -0.064 & \colorbox[HTML]{F9CBD0}{-0.180} &  -0.019 & -0.113 &  \colorbox[HTML]{F9CBD0}{-0.295} & hallucin. &-1.560 & -2.055\\
         &whole & -0.040 & -0.103 & \colorbox[HTML]{F9CBD0}{-0.347} &  -0.079 &  -0.173 & \colorbox[HTML]{F9CBD0}{-0.509} &\\ \addlinespace
         \textbf{average} & & \textbf{-0.037} & \textbf{-0.068} & \colorbox[HTML]{F9CBD0}{\textbf{-0.232}} & \textbf{-0.068} & \textbf{-0.123} & \colorbox[HTML]{F9CBD0}{\textbf{-0.360}}
         
    \end{tabular}
    \caption{Effects of randomly adding, substituting or deleting a digit in a number or a letter in a noun or named entity (NE) compared to the controls (alternative translation, copy of the source or hallucination). The numbers show the average difference to the MBR score for the reference (left) and 1-best beam search output (right) when using \texttt{wmt20-comet-da} as the utility function. Red means the sensitivity for random nouns is larger than for both numbers and named entities.}
    \label{tab:sensitivity}
\end{table*}

\section{MBR-Based Sensitivity Analysis}
\label{sec:sensitivity}
Our findings in the previous section stand in contrast to the corrupted reference analysis performed as part of the WMT 2021 metrics shared task \citep{freitag-etal-2021-results} where COMET mostly preferred the correct alternative human translation to one with swapped numbers when comparing to the reference. In reality, we will seldom have a hypothesis pool with a perfect translation and variants of it that only differ in one aspect. Ideally, evaluation metrics should be able to order translation hypotheses with many different error types according to their severity. Therefore, it makes sense to compare how much metrics punish different error types.

Since our previous analysis showed that many samples with number and named entity mismatches are chosen in MBR decoding, this indicates that COMET is not as sensitive to these error types as to other errors. To further support this finding, we propose to look more closely at how COMET behaves with different error types. As described in Section \ref{sec:mbr}, in MBR decoding, every candidate translation is assigned a score that represents the average similarity to the support hypotheses. Consequently, if the support is kept constant and a targeted change is made to a candidate translation, the  difference in this MBR score indicates how sensitive the utility function was towards this change. We term this an ``MBR-based sensitivity analysis''.

To measure COMET's sensitivity towards changes in numbers and named entities,
we create a candidate pool that consists of the reference translation and several changed variants. Note that the support still contains the same 100 samples that were used to find the MBR-decoded outputs described in Section \ref{sec:exploration}. 
In particular, we make the following targeted changes to the reference to measure the sensitivity towards each change:
\begin{itemize}
    \item \textbf{num\textsubscript{add}}: one digit is added to a number at a random position.
    \item \textbf{num\textsubscript{del}}: one digit is removed from a number at a random position.
    \item \textbf{num\textsubscript{sub}}: one digit is substituted with another digit in a number  at a random position.
    \item \textbf{num\textsubscript{whole}}: one entire number is substituted with another number.
    \item \textbf{NE\textsubscript{add}}: one letter is added to a named entity at a random position.
    \item \textbf{NE\textsubscript{del}}: one letter is removed from a named entity at a random position.
    \item \textbf{NE\textsubscript{sub}}: one letter is substituted with another letter in a named entity at a random position.
    \item \textbf{NE\textsubscript{whole}}: a named entity is substituted with another named entity.
\end{itemize}

As reference points, we also apply the same types of changes to random nouns in the reference:

\begin{itemize}
    \item \textbf{noun\textsubscript{add}}: one letter is added to a random noun at a random position.
    \item \textbf{noun\textsubscript{del}}: one letter is removed from a random noun at a random position.
    \item \textbf{noun\textsubscript{sub}}: one letter is substituted with another letter in a random noun at a random position.
    \item \textbf{noun\textsubscript{whole}}: a random noun is substituted with another noun.
\end{itemize}

Additionally, our candidate pool contains the following hypotheses to be used as controls:

\begin{itemize}
    \item \textbf{alternative}: the second human reference provided as part of the WMT 2021 news shared task simulating an alternative translation.
    \item \textbf{copy}: the original, unchanged source sentence simulating a model that simply copied the source to the decoder side.
    \item \textbf{hallucination}: a sentence that is completely unrelated to the source and randomly picked from a larger corpus.
 \end{itemize}   

We use the same tools to identify numbers and named entities as in Section \ref{sec:exploration} to create these perturbations of the reference. For each newly created candidate, we compute the difference to the MBR score of the reference. We then average those differences across sentences for each perturbation type. The results for the sensitivity analysis with the \texttt{wmt20-comet-da} model can be seen in the left part of Table \ref{tab:sensitivity}. We focus here on the \texttt{wmt20-comet-da} model since this is currently the model the authors recommend to use.\footnote{\url{https://github.com/Unbabel/COMET/blob/master/METRICS.md}}

The controls, i.e. alternative translation, copied sentence and hallucination, behave as expected. The MBR score difference to the hallucination is by far the largest, followed by the copied source. For the alternative reference, we see the smallest MBR score difference.\footnote{Note that this is due to averaging over sentences where the alternative sometimes gets a higher, sometimes a lower score. The average absolute difference is 0.111 which shows that the difference to the alternative of an individual sentence can be much larger.} More importantly, all targeted changes to \emph{numbers} or \emph{named entities} result in a much smaller difference in MBR score compared to changes to the \emph{random nouns}. This shows that COMET is not as sensitive to such discrepancies as it should be since such mistranslations can drastically alter the meaning. Both BLEU and chrF++ are more sensitive to changes to numbers and named entities than to random nouns (see Appendix \ref{app:sensitivity_bleu}).

Following our discussion of error sources at the beginning of Section \ref{sec:exploration}, it is a valid concern that if we were to compare the candidates to high-quality support translations rather than samples, COMET may be more sensitive toward number and named entity differences as there would be fewer other discrepancies between the candidates and the support. To test if this is the case, we repeat the sensitivity analysis but now use the two alternative references as the support instead of the 100 samples that were used before. The candidates are formed by applying the same perturbations as before to the 1-best beam search output instead of the reference. This mimics an oracle setup. The results for this experiment are shown in the middle of Table \ref{tab:sensitivity}. Note that we cannot compare to an alternative translation for the beam search output in this setup. 

The differences in the MBR score of the unperturbed beam search output are generally larger in this setup, which indicates that COMET is indeed more sensitive to errors when used as intended, i.e.\ with high-quality translations and correct references. 
However, we can still see that the perturbations made to random nouns result in much larger differences than perturbations made to numbers or named entities.
This indicates that the problem cannot be attributed to the MBR decoding setting and low-quality pseudo-references alone.

\begin{figure*}[htpb]
    \centering
    \includegraphics[width=\textwidth]{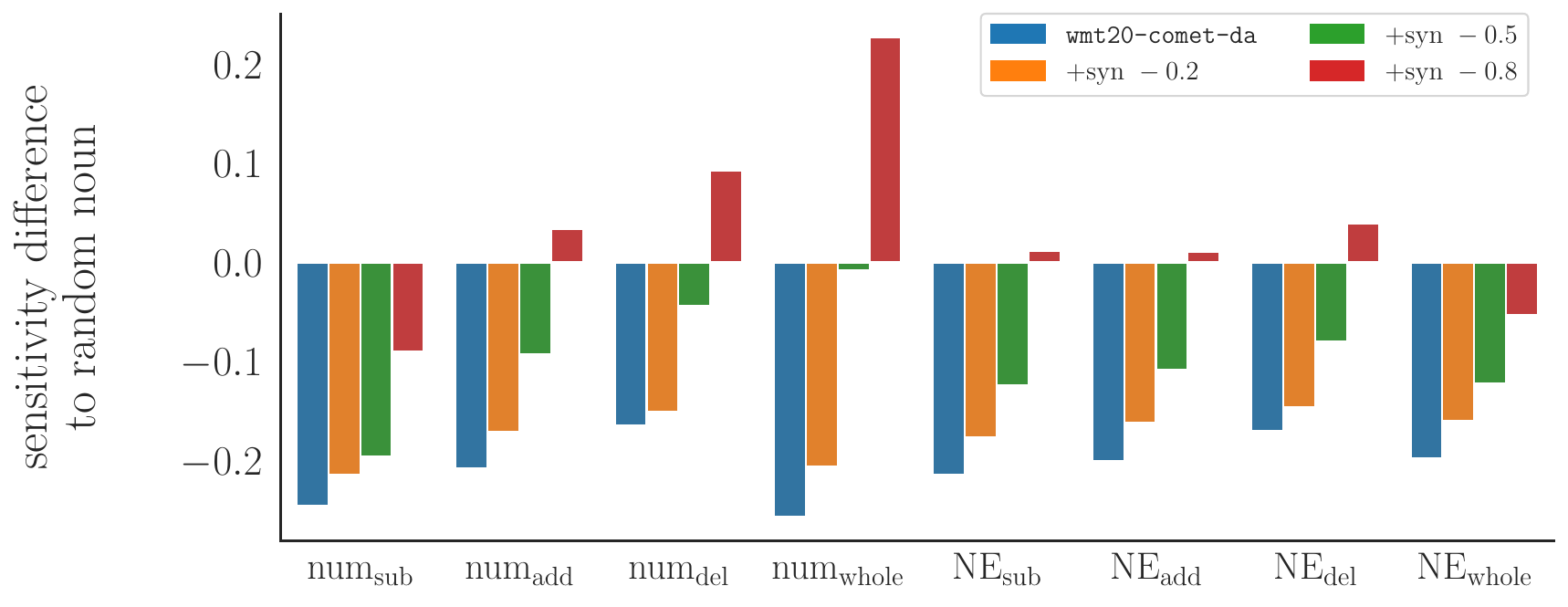}
    \caption{Difference in sensitivity to the same error type applied to a random noun for the de-en test set with samples as support. Comparing the original \texttt{wmt20-comet-da} to three retrained models, with different amounts subtracted from the original score for synthetic examples (-0.2, -0.5 and -0.8).}
    \label{fig:diff_nouns}
\end{figure*}

\section{COMET Retraining}
\label{sec:retrain}

One possible explanation for the low sensitivity of COMET to perturbations of numbers and named entities is that these errors are too rare in the WMT outputs used to train COMET.
 We decide to retrain COMET on the original training data plus added synthetic data on which we perform the same perturbations as described in Section \ref{sec:sensitivity}. The idea is that the newly trained model is more sensitive toward named entity or number mismatches between the translation and its reference and/or source.

To retrain the \texttt{wmt20-comet-da} model, we use the data from the WMT metrics shared tasks collected in the years 2017 to 2019 \citep{bojar-etal-2017-results, ma-etal-2018-results, ma-etal-2019-results} as training data. For every de$\rightarrow$en or en$\rightarrow$de system output that contains a number or a named entity, we randomly apply one of the perturbations described in Section \ref{sec:sensitivity} (except for the perturbations of random nouns and whole named entities). To encourage COMET to punish such synthetically inserted mismatches, we modify the scores of the original examples by subtracting a penalty from the z-score of the Direct Assessment (DA) score. We retrain three different models with penalties of -0.2, -0.5 and -0.8 respectively. Within every experiment, the penalty is the same for all error classes. The resulting $\sim$61k synthetic training examples are then added to the $\sim$640k original examples which means that roughly 10\% of the data are synthetic.\footnote{We also trained models with larger amounts of synthetic data but did not see an improvement (see Appendix \ref{app:synthetic}).}

We follow the hyperparameter suggestions in \citet{rei-etal-2020-unbabels} for retraining COMET but we do not perform model averaging. The models are trained for two epochs and the hyperparameters are listed in Appendix \ref{app:retrain}.
We ensure that the retrained models still perform as well as the original model on the WMT 2020 metrics shared task \cite{mathur-etal-2020-results}. The average difference in system-level Pearson correlation to the original COMET model lies within 0.006 for all three penalties. The full results can be found in Appendix \ref{app:corr}.

The effects of retraining with different penalties can be seen in Figure \ref{fig:diff_nouns} (tables in Appendix \ref{app:punish}). Subtracting -0.2 from the original scores for synthetic examples can slightly reduce the difference between the MBR scores for numbers / named entities and random nouns with the same error types. Retraining with -0.5 subtracted from the original score improves this further but still cannot close this gap completely. With a penalty of -0.8, we now see a larger sensitivity to numbers and named entities than to random nouns for several error types. However, the difference to random nouns is still rather high for substituting a digit in numbers.

When repeating the automatic analysis from Section \ref{sec:exploration} with the penalty -0.8 model, we see that retraining does improve the F1-scores (see last row in Table \ref{tab:automatic}). However, the retrained COMET model can still not beat non-neural utility functions which indicates that it is still less sensitive to mismatches in numbers and named entities.

From this experiment, we conclude that removing such blind spots from COMET - once identified - might need more effort than simply training on additional synthetic data. We hypothesise is that the XLM-R component learns very similar representations for numbers and rare words like named entities during pretraining which could be hard to reverse with finetuning only. \citet{lin-etal-2020-birds} show that pretrained language models are surprisingly bad at guessing the correct number from context (e.g. "A bird usually has [MASK] legs.") which supports this hypothesis. Several other works also find that task-specific models often struggle with numbers and named entities such as in summarisation \citep{zhao-etal-2020-reducing} or question answering \citep{dua-etal-2019-drop, kim-etal-2021-seen}. We leave a more extensive analysis of biases in the human evaluation training data (e.g. unpunished number mismatches) and further experiments on weakness-targeted training for future work.

\section{Conclusion}
Identifying weaknesses of neural machine translation evaluation metrics becomes more important as these essentially ``black box'' evaluation tools become more popular and are optimised towards during model development. We show that MBR decoding can be used to explore biases of such metrics. Through a case study, we show that COMET is relatively insensitive to mistranslated numbers and named entities. This can be seen both in the MBR-decoded output which contains a higher number of these errors compared to beam search (or MBR with other utility functions) and in an MBR-based sensitivity analysis which compares the differences in MBR scores that arise when such errors are introduced to a candidate translation.
We also show that this insensitivity is not simply the result of insufficient training data containing such errors: retraining COMET with additional synthetic data did not fully alleviate this weakness.

While errors related to number and named entity translation were very salient in our exploration, we do not claim that this case study is exhaustive. In our manual analysis, we also see anecdotal evidence of polarity errors and nonsensical German compounds. We hope our findings motivate further research into identifying and mitigating biases of neural machine translation metrics -- we envision that actively searching for biases in neural metrics, for example by using them as utility functions in MBR, could become an important step during metric development.

\section*{Acknowledgements}
We are grateful to Ricardo Rei for the support in retraining COMET and the anonymous reviewers for their helpful feedback. We also thank Janis Goldzycher, Mathias Müller, Nikita Moghe and Tannon Kew for their valuable inputs and interesting discussions throughout the development of this paper. This work was funded by the Swiss National Science Foundation (project MUTAMUR; no. 176727) and made use of the infrastructure services provided by S3IT, the Service and Support for Science IT team at the University of Zurich.

\section*{Ethical Considerations}
In our work, we only use publicly available toolkits and datasets and do not collect any additional data. Our experiments also do not involve human annotators (other than ourselves). The main contribution of our paper is a new approach for identifying weaknesses in neural machine translation evaluation metrics using MBR decoding. We believe this approach is largely beneficial to the research community as a tool to investigate 
``blind spots'' of metrics and we do not see any immediate risks.

\section*{Limitations}
We limited our analysis in this work to the en$\leftrightarrow$de translation directions and one machine translation evaluation metric, namely COMET. Consequently, we cannot draw any conclusions on whether the identified weaknesses are specific to COMET or also apply to other neural machine translation evaluation metrics and language pairs. We leave such exploration for future work. While our approach for identifying weaknesses in evaluation metrics is readily applicable to other surface-level or neural metrics, the runtime for MBR decoding can explode if the similarity computation cannot be parallelised or the size of the sample pool is increased. However, since our proposed approach is a tool for metric analysis and is not intended to be run regularly, we believe an increased runtime is not obstructive. 

Another limitation is that we do not use a state-of-the-art machine translation model (in terms of data size) to generate the samples for our metric analysis. This does, however, not limit our findings that COMET is not as sensitive to number and named entity differences as it should be. Even if machine translation models may produce fewer mistakes of this nature in the future, eliminating such weaknesses remains relevant, for example, if COMET is used for Minimum Risk Training.

Finally, while our experiments indicate that weaknesses related to number and named entity changes cannot easily be eliminated by retraining on synthetic data, alternative strategies to create or retrain on synthetic data may be more successful.

\bibliography{anthology,custom}

\begin{thebibliography}{48}
\expandafter\ifx\csname natexlab\endcsname\relax\def\natexlab#1{#1}\fi

\bibitem[{Akhbardeh et~al.(2021)Akhbardeh, Arkhangorodsky, Biesialska, Bojar,
  Chatterjee, Chaudhary, Costa-jussa, Espa{\~n}a-Bonet, Fan, Federmann,
  Freitag, Graham, Grundkiewicz, Haddow, Harter, Heafield, Homan, Huck,
  Amponsah-Kaakyire, Kasai, Khashabi, Knight, Kocmi, Koehn, Lourie, Monz,
  Morishita, Nagata, Nagesh, Nakazawa, Negri, Pal, Tapo, Turchi, Vydrin, and
  Zampieri}]{akhbardeh-etal-2021-findings}
Farhad Akhbardeh, Arkady Arkhangorodsky, Magdalena Biesialska, Ond{\v{r}}ej
  Bojar, Rajen Chatterjee, Vishrav Chaudhary, Marta~R. Costa-jussa, Cristina
  Espa{\~n}a-Bonet, Angela Fan, Christian Federmann, Markus Freitag, Yvette
  Graham, Roman Grundkiewicz, Barry Haddow, Leonie Harter, Kenneth Heafield,
  Christopher Homan, Matthias Huck, Kwabena Amponsah-Kaakyire, Jungo Kasai,
  Daniel Khashabi, Kevin Knight, Tom Kocmi, Philipp Koehn, Nicholas Lourie,
  Christof Monz, Makoto Morishita, Masaaki Nagata, Ajay Nagesh, Toshiaki
  Nakazawa, Matteo Negri, Santanu Pal, Allahsera~Auguste Tapo, Marco Turchi,
  Valentin Vydrin, and Marcos Zampieri. 2021.
\newblock \href {https://aclanthology.org/2021.wmt-1.1} {Findings of the 2021
  conference on machine translation ({WMT}21)}.
\newblock In \emph{Proceedings of the Sixth Conference on Machine Translation},
  pages 1--88, Online. Association for Computational Linguistics.

\bibitem[{Banerjee and Lavie(2005)}]{banerjee-lavie-2005-meteor}
Satanjeev Banerjee and Alon Lavie. 2005.
\newblock \href {https://aclanthology.org/W05-0909} {{METEOR}: An automatic
  metric for {MT} evaluation with improved correlation with human judgments}.
\newblock In \emph{Proceedings of the {ACL} Workshop on Intrinsic and Extrinsic
  Evaluation Measures for Machine Translation and/or Summarization}, pages
  65--72, Ann Arbor, Michigan. Association for Computational Linguistics.

\bibitem[{Bojar et~al.(2018)Bojar, Federmann, Fishel, Graham, Haddow, Huck,
  Koehn, and Monz}]{bojar-etal-2018-findings}
Ond{\v{r}}ej Bojar, Christian Federmann, Mark Fishel, Yvette Graham, Barry
  Haddow, Matthias Huck, Philipp Koehn, and Christof Monz. 2018.
\newblock \href {https://doi.org/10.18653/v1/W18-6401} {Findings of the 2018
  conference on machine translation ({WMT}18)}.
\newblock In \emph{Proceedings of the Third Conference on Machine Translation:
  Shared Task Papers}, pages 272--303, Belgium, Brussels. Association for
  Computational Linguistics.

\bibitem[{Bojar et~al.(2017)Bojar, Graham, and
  Kamran}]{bojar-etal-2017-results}
Ond{\v{r}}ej Bojar, Yvette Graham, and Amir Kamran. 2017.
\newblock \href {https://doi.org/10.18653/v1/W17-4755} {Results of the {WMT}17
  metrics shared task}.
\newblock In \emph{Proceedings of the Second Conference on Machine
  Translation}, pages 489--513, Copenhagen, Denmark. Association for
  Computational Linguistics.

\bibitem[{Conneau et~al.(2020)Conneau, Khandelwal, Goyal, Chaudhary, Wenzek,
  Guzm{\'a}n, Grave, Ott, Zettlemoyer, and
  Stoyanov}]{conneau-etal-2020-unsupervised}
Alexis Conneau, Kartikay Khandelwal, Naman Goyal, Vishrav Chaudhary, Guillaume
  Wenzek, Francisco Guzm{\'a}n, Edouard Grave, Myle Ott, Luke Zettlemoyer, and
  Veselin Stoyanov. 2020.
\newblock \href {https://doi.org/10.18653/v1/2020.acl-main.747} {Unsupervised
  cross-lingual representation learning at scale}.
\newblock In \emph{Proceedings of the 58th Annual Meeting of the Association
  for Computational Linguistics}, pages 8440--8451, Online. Association for
  Computational Linguistics.

\bibitem[{Dua et~al.(2019)Dua, Wang, Dasigi, Stanovsky, Singh, and
  Gardner}]{dua-etal-2019-drop}
Dheeru Dua, Yizhong Wang, Pradeep Dasigi, Gabriel Stanovsky, Sameer Singh, and
  Matt Gardner. 2019.
\newblock \href {https://doi.org/10.18653/v1/N19-1246} {{DROP}: A reading
  comprehension benchmark requiring discrete reasoning over paragraphs}.
\newblock In \emph{Proceedings of the 2019 Conference of the North {A}merican
  Chapter of the Association for Computational Linguistics: Human Language
  Technologies, Volume 1 (Long and Short Papers)}, pages 2368--2378,
  Minneapolis, Minnesota. Association for Computational Linguistics.

\bibitem[{Eikema and Aziz(2020)}]{eikema-aziz-2020-map}
Bryan Eikema and Wilker Aziz. 2020.
\newblock \href {https://doi.org/10.18653/v1/2020.coling-main.398} {Is {MAP}
  decoding all you need? the inadequacy of the mode in neural machine
  translation}.
\newblock In \emph{Proceedings of the 28th International Conference on
  Computational Linguistics}, pages 4506--4520, Barcelona, Spain (Online).
  International Committee on Computational Linguistics.

\bibitem[{Eikema and Aziz(2021)}]{eikema2021samplingbased}
Bryan Eikema and Wilker Aziz. 2021.
\newblock \href {http://arxiv.org/abs/2108.04718} {Sampling-based minimum bayes
  risk decoding for neural machine translation}.

\bibitem[{Fomicheva et~al.(2020)Fomicheva, Specia, and
  Guzm{\'a}n}]{fomicheva-etal-2020-multi}
Marina Fomicheva, Lucia Specia, and Francisco Guzm{\'a}n. 2020.
\newblock \href {https://doi.org/10.18653/v1/2020.acl-main.113}
  {Multi-hypothesis machine translation evaluation}.
\newblock In \emph{Proceedings of the 58th Annual Meeting of the Association
  for Computational Linguistics}, pages 1218--1232, Online. Association for
  Computational Linguistics.

\bibitem[{Freitag et~al.(2021{\natexlab{a}})Freitag, Foster, Grangier,
  Ratnakar, Tan, and Macherey}]{freitag-etal-2021-experts}
Markus Freitag, George Foster, David Grangier, Viresh Ratnakar, Qijun Tan, and
  Wolfgang Macherey. 2021{\natexlab{a}}.
\newblock \href {https://doi.org/10.1162/tacl_a_00437} {Experts, errors, and
  context: A large-scale study of human evaluation for machine translation}.
\newblock \emph{Transactions of the Association for Computational Linguistics},
  9:1460--1474.

\bibitem[{Freitag et~al.(2022)Freitag, Grangier, Tan, and
  Liang}]{freitag-etal-2022-high}
Markus Freitag, David Grangier, Qijun Tan, and Bowen Liang. 2022.
\newblock \href {https://doi.org/10.1162/tacl_a_00491} {High quality rather
  than high model probability: Minimum {B}ayes risk decoding with neural
  metrics}.
\newblock \emph{Transactions of the Association for Computational Linguistics},
  10:811--825.

\bibitem[{Freitag et~al.(2021{\natexlab{b}})Freitag, Rei, Mathur, Lo, Stewart,
  Foster, Lavie, and Bojar}]{freitag-etal-2021-results}
Markus Freitag, Ricardo Rei, Nitika Mathur, Chi-kiu Lo, Craig Stewart, George
  Foster, Alon Lavie, and Ond{\v{r}}ej Bojar. 2021{\natexlab{b}}.
\newblock \href {https://aclanthology.org/2021.wmt-1.73} {Results of the
  {WMT}21 metrics shared task: Evaluating metrics with expert-based human
  evaluations on {TED} and news domain}.
\newblock In \emph{Proceedings of the Sixth Conference on Machine Translation},
  pages 733--774, Online. Association for Computational Linguistics.

\bibitem[{Goel and Byrne(2000)}]{GOEL2000115}
Vaibhava Goel and William~J Byrne. 2000.
\newblock \href {https://doi.org/https://doi.org/10.1006/csla.2000.0138}
  {Minimum bayes-risk automatic speech recognition}.
\newblock \emph{Computer Speech \& Language}, 14(2):115--135.

\bibitem[{Graham et~al.(2017)Graham, Baldwin, Moffat, and
  Zobel}]{graham_baldwin_moffat_zobel_2017}
Yvette Graham, Timothy Baldwin, Alistair Moffat, and Justin Zobel. 2017.
\newblock \href {https://doi.org/10.1017/S1351324915000339} {Can machine
  translation systems be evaluated by the crowd alone}.
\newblock \emph{Natural Language Engineering}, 23(1):3–30.

\bibitem[{Hanna and Bojar(2021)}]{hanna-bojar-2021-fine}
Michael Hanna and Ond{\v{r}}ej Bojar. 2021.
\newblock \href {https://aclanthology.org/2021.wmt-1.59} {A fine-grained
  analysis of {BERTS}core}.
\newblock In \emph{Proceedings of the Sixth Conference on Machine Translation},
  pages 507--517, Online. Association for Computational Linguistics.

\bibitem[{Honnibal et~al.(2020)Honnibal, Montani, Van~Landeghem, and
  Boyd}]{spacy}
Matthew Honnibal, Ines Montani, Sofie Van~Landeghem, and Adriane Boyd. 2020.
\newblock \href {https://doi.org/10.5281/zenodo.1212303} {{spaCy:
  Industrial-strength Natural Language Processing in Python}}.

\bibitem[{Junczys-Dowmunt et~al.(2018)Junczys-Dowmunt, Grundkiewicz, Dwojak,
  Hoang, Heafield, Neckermann, Seide, Germann, Aji, Bogoychev, Martins, and
  Birch}]{junczys-dowmunt-etal-2018-marian}
Marcin Junczys-Dowmunt, Roman Grundkiewicz, Tomasz Dwojak, Hieu Hoang, Kenneth
  Heafield, Tom Neckermann, Frank Seide, Ulrich Germann, Alham~Fikri Aji,
  Nikolay Bogoychev, Andr{\'e} F.~T. Martins, and Alexandra Birch. 2018.
\newblock \href {https://doi.org/10.18653/v1/P18-4020} {{M}arian: Fast neural
  machine translation in {C}++}.
\newblock In \emph{Proceedings of {ACL} 2018, System Demonstrations}, pages
  116--121, Melbourne, Australia. Association for Computational Linguistics.

\bibitem[{Kim et~al.(2021)Kim, Hong, Kim, Kang, and
  Myaeng}]{kim-etal-2021-seen}
Jeonghwan Kim, Giwon Hong, Kyung-min Kim, Junmo Kang, and Sung-Hyon Myaeng.
  2021.
\newblock \href {https://doi.org/10.18653/v1/2021.emnlp-main.563} {Have you
  seen that number? investigating extrapolation in question answering models}.
\newblock In \emph{Proceedings of the 2021 Conference on Empirical Methods in
  Natural Language Processing}, pages 7031--7037, Online and Punta Cana,
  Dominican Republic. Association for Computational Linguistics.

\bibitem[{Kingma and Ba(2015)}]{DBLP:journals/corr/KingmaB14}
Diederik~P. Kingma and Jimmy Ba. 2015.
\newblock \href {http://arxiv.org/abs/1412.6980} {{Adam: {A} Method for
  Stochastic Optimization}}.
\newblock In \emph{3rd International Conference on Learning Representations,
  {ICLR} 2015, San Diego, CA, USA, May 7-9, 2015, Conference Track
  Proceedings}.

\bibitem[{Kocmi et~al.(2021)Kocmi, Federmann, Grundkiewicz, Junczys-Dowmunt,
  Matsushita, and Menezes}]{kocmi-etal-2021-ship}
Tom Kocmi, Christian Federmann, Roman Grundkiewicz, Marcin Junczys-Dowmunt,
  Hitokazu Matsushita, and Arul Menezes. 2021.
\newblock \href {https://aclanthology.org/2021.wmt-1.57} {To ship or not to
  ship: An extensive evaluation of automatic metrics for machine translation}.
\newblock In \emph{Proceedings of the Sixth Conference on Machine Translation},
  pages 478--494, Online. Association for Computational Linguistics.

\bibitem[{Kudo and Richardson(2018)}]{kudo-richardson-2018-sentencepiece}
Taku Kudo and John Richardson. 2018.
\newblock \href {https://doi.org/10.18653/v1/D18-2012} {{S}entence{P}iece: A
  simple and language independent subword tokenizer and detokenizer for neural
  text processing}.
\newblock In \emph{Proceedings of the 2018 Conference on Empirical Methods in
  Natural Language Processing: System Demonstrations}, pages 66--71, Brussels,
  Belgium. Association for Computational Linguistics.

\bibitem[{Kumar and Byrne(2004)}]{kumar-byrne-2004-minimum}
Shankar Kumar and William Byrne. 2004.
\newblock \href {https://aclanthology.org/N04-1022} {Minimum {B}ayes-risk
  decoding for statistical machine translation}.
\newblock In \emph{Proceedings of the Human Language Technology Conference of
  the North {A}merican Chapter of the Association for Computational
  Linguistics: {HLT}-{NAACL} 2004}, pages 169--176, Boston, Massachusetts, USA.
  Association for Computational Linguistics.

\bibitem[{Lin et~al.(2020)Lin, Lee, Khanna, and Ren}]{lin-etal-2020-birds}
Bill~Yuchen Lin, Seyeon Lee, Rahul Khanna, and Xiang Ren. 2020.
\newblock \href {https://doi.org/10.18653/v1/2020.emnlp-main.557} {{B}irds have
  four legs?! {N}umer{S}ense: {P}robing {N}umerical {C}ommonsense {K}nowledge
  of {P}re-{T}rained {L}anguage {M}odels}.
\newblock In \emph{Proceedings of the 2020 Conference on Empirical Methods in
  Natural Language Processing (EMNLP)}, pages 6862--6868, Online. Association
  for Computational Linguistics.

\bibitem[{Lo(2020)}]{lo-2020-extended}
Chi-kiu Lo. 2020.
\newblock \href {https://aclanthology.org/2020.wmt-1.99} {Extended study on
  using pretrained language models and {Y}i{S}i-1 for machine translation
  evaluation}.
\newblock In \emph{Proceedings of the Fifth Conference on Machine Translation},
  pages 895--902, Online. Association for Computational Linguistics.

\bibitem[{Ma et~al.(2018)Ma, Bojar, and Graham}]{ma-etal-2018-results}
Qingsong Ma, Ond{\v{r}}ej Bojar, and Yvette Graham. 2018.
\newblock \href {https://doi.org/10.18653/v1/W18-6450} {Results of the {WMT}18
  metrics shared task: Both characters and embeddings achieve good
  performance}.
\newblock In \emph{Proceedings of the Third Conference on Machine Translation:
  Shared Task Papers}, pages 671--688, Belgium, Brussels. Association for
  Computational Linguistics.

\bibitem[{Ma et~al.(2019)Ma, Wei, Bojar, and Graham}]{ma-etal-2019-results}
Qingsong Ma, Johnny Wei, Ond{\v{r}}ej Bojar, and Yvette Graham. 2019.
\newblock \href {https://doi.org/10.18653/v1/W19-5302} {Results of the {WMT}19
  metrics shared task: Segment-level and strong {MT} systems pose big
  challenges}.
\newblock In \emph{Proceedings of the Fourth Conference on Machine Translation
  (Volume 2: Shared Task Papers, Day 1)}, pages 62--90, Florence, Italy.
  Association for Computational Linguistics.

\bibitem[{Mathur et~al.(2020)Mathur, Wei, Freitag, Ma, and
  Bojar}]{mathur-etal-2020-results}
Nitika Mathur, Johnny Wei, Markus Freitag, Qingsong Ma, and Ond{\v{r}}ej Bojar.
  2020.
\newblock \href {https://aclanthology.org/2020.wmt-1.77} {Results of the
  {WMT}20 metrics shared task}.
\newblock In \emph{Proceedings of the Fifth Conference on Machine Translation},
  pages 688--725, Online. Association for Computational Linguistics.

\bibitem[{M{\"u}ller and Sennrich(2021)}]{muller-sennrich-2021-understanding}
Mathias M{\"u}ller and Rico Sennrich. 2021.
\newblock \href {https://doi.org/10.18653/v1/2021.acl-long.22} {Understanding
  the properties of minimum {B}ayes risk decoding in neural machine
  translation}.
\newblock In \emph{Proceedings of the 59th Annual Meeting of the Association
  for Computational Linguistics and the 11th International Joint Conference on
  Natural Language Processing (Volume 1: Long Papers)}, pages 259--272, Online.
  Association for Computational Linguistics.

\bibitem[{Nakov et~al.(2012)Nakov, Guzman, and
  Vogel}]{nakov-etal-2012-optimizing}
Preslav Nakov, Francisco Guzman, and Stephan Vogel. 2012.
\newblock \href {https://aclanthology.org/C12-1121} {Optimizing for
  sentence-level {BLEU}+1 yields short translations}.
\newblock In \emph{Proceedings of {COLING} 2012}, pages 1979--1994, Mumbai,
  India. The COLING 2012 Organizing Committee.

\bibitem[{Papineni et~al.(2002)Papineni, Roukos, Ward, and
  Zhu}]{papineni-etal-2002-bleu}
Kishore Papineni, Salim Roukos, Todd Ward, and Wei-Jing Zhu. 2002.
\newblock \href {https://doi.org/10.3115/1073083.1073135} {{B}leu: a method for
  automatic evaluation of machine translation}.
\newblock In \emph{Proceedings of the 40th Annual Meeting of the Association
  for Computational Linguistics}, pages 311--318, Philadelphia, Pennsylvania,
  USA. Association for Computational Linguistics.

\bibitem[{Popovi{\'c}(2015)}]{popovic-2015-chrf}
Maja Popovi{\'c}. 2015.
\newblock \href {https://doi.org/10.18653/v1/W15-3049} {chr{F}: character
  n-gram {F}-score for automatic {MT} evaluation}.
\newblock In \emph{Proceedings of the Tenth Workshop on Statistical Machine
  Translation}, pages 392--395, Lisbon, Portugal. Association for Computational
  Linguistics.

\bibitem[{Popovi{\'c}(2017)}]{popovic-2017-chrf}
Maja Popovi{\'c}. 2017.
\newblock \href {https://doi.org/10.18653/v1/W17-4770} {chr{F}++: words helping
  character n-grams}.
\newblock In \emph{Proceedings of the Second Conference on Machine
  Translation}, pages 612--618, Copenhagen, Denmark. Association for
  Computational Linguistics.

\bibitem[{Post(2018)}]{post-2018-call}
Matt Post. 2018.
\newblock \href {https://doi.org/10.18653/v1/W18-6319} {A call for clarity in
  reporting {BLEU} scores}.
\newblock In \emph{Proceedings of the Third Conference on Machine Translation:
  Research Papers}, pages 186--191, Brussels, Belgium. Association for
  Computational Linguistics.

\bibitem[{Press and Wolf(2017)}]{press-wolf-2017-using}
Ofir Press and Lior Wolf. 2017.
\newblock \href {https://aclanthology.org/E17-2025} {Using the output embedding
  to improve language models}.
\newblock In \emph{Proceedings of the 15th Conference of the {E}uropean Chapter
  of the Association for Computational Linguistics: Volume 2, Short Papers},
  pages 157--163, Valencia, Spain. Association for Computational Linguistics.

\bibitem[{Provilkov et~al.(2020)Provilkov, Emelianenko, and
  Voita}]{provilkov-etal-2020-bpe}
Ivan Provilkov, Dmitrii Emelianenko, and Elena Voita. 2020.
\newblock \href {https://doi.org/10.18653/v1/2020.acl-main.170} {{BPE}-dropout:
  Simple and effective subword regularization}.
\newblock In \emph{Proceedings of the 58th Annual Meeting of the Association
  for Computational Linguistics}, pages 1882--1892, Online. Association for
  Computational Linguistics.

\bibitem[{Rei et~al.(2021)Rei, Farinha, Zerva, van Stigt, Stewart, Ramos,
  Glushkova, Martins, and Lavie}]{rei-etal-2021-references}
Ricardo Rei, Ana~C Farinha, Chrysoula Zerva, Daan van Stigt, Craig Stewart,
  Pedro Ramos, Taisiya Glushkova, Andr{\'e} F.~T. Martins, and Alon Lavie.
  2021.
\newblock \href {https://aclanthology.org/2021.wmt-1.111} {Are references
  really needed? unbabel-{IST} 2021 submission for the metrics shared task}.
\newblock In \emph{Proceedings of the Sixth Conference on Machine Translation},
  pages 1030--1040, Online. Association for Computational Linguistics.

\bibitem[{Rei et~al.(2020{\natexlab{a}})Rei, Stewart, Farinha, and
  Lavie}]{rei-etal-2020-comet}
Ricardo Rei, Craig Stewart, Ana~C Farinha, and Alon Lavie. 2020{\natexlab{a}}.
\newblock \href {https://doi.org/10.18653/v1/2020.emnlp-main.213} {{COMET}: A
  neural framework for {MT} evaluation}.
\newblock In \emph{Proceedings of the 2020 Conference on Empirical Methods in
  Natural Language Processing (EMNLP)}, pages 2685--2702, Online. Association
  for Computational Linguistics.

\bibitem[{Rei et~al.(2020{\natexlab{b}})Rei, Stewart, Farinha, and
  Lavie}]{rei-etal-2020-unbabels}
Ricardo Rei, Craig Stewart, Ana~C Farinha, and Alon Lavie. 2020{\natexlab{b}}.
\newblock \href {https://aclanthology.org/2020.wmt-1.101} {Unbabel{'}s
  participation in the {WMT}20 metrics shared task}.
\newblock In \emph{Proceedings of the Fifth Conference on Machine Translation},
  pages 911--920, Online. Association for Computational Linguistics.

\bibitem[{Sellam et~al.(2020)Sellam, Das, and Parikh}]{sellam-etal-2020-bleurt}
Thibault Sellam, Dipanjan Das, and Ankur Parikh. 2020.
\newblock \href {https://doi.org/10.18653/v1/2020.acl-main.704} {{BLEURT}:
  Learning robust metrics for text generation}.
\newblock In \emph{Proceedings of the 58th Annual Meeting of the Association
  for Computational Linguistics}, pages 7881--7892, Online. Association for
  Computational Linguistics.

\bibitem[{Sennrich et~al.(2017)Sennrich, Firat, Cho, Birch, Haddow, Hitschler,
  Junczys-Dowmunt, L{\"a}ubli, Miceli~Barone, Mokry, and
  N{\u{a}}dejde}]{sennrich-etal-2017-nematus}
Rico Sennrich, Orhan Firat, Kyunghyun Cho, Alexandra Birch, Barry Haddow,
  Julian Hitschler, Marcin Junczys-Dowmunt, Samuel L{\"a}ubli, Antonio~Valerio
  Miceli~Barone, Jozef Mokry, and Maria N{\u{a}}dejde. 2017.
\newblock \href {https://aclanthology.org/E17-3017} {{N}ematus: a toolkit for
  neural machine translation}.
\newblock In \emph{Proceedings of the Software Demonstrations of the 15th
  Conference of the {E}uropean Chapter of the Association for Computational
  Linguistics}, pages 65--68, Valencia, Spain. Association for Computational
  Linguistics.

\bibitem[{Sennrich et~al.(2016)Sennrich, Haddow, and
  Birch}]{sennrich-etal-2016-neural}
Rico Sennrich, Barry Haddow, and Alexandra Birch. 2016.
\newblock \href {https://doi.org/10.18653/v1/P16-1162} {Neural machine
  translation of rare words with subword units}.
\newblock In \emph{Proceedings of the 54th Annual Meeting of the Association
  for Computational Linguistics (Volume 1: Long Papers)}, pages 1715--1725,
  Berlin, Germany. Association for Computational Linguistics.

\bibitem[{Shen et~al.(2016)Shen, Cheng, He, He, Wu, Sun, and
  Liu}]{shen-etal-2016-minimum}
Shiqi Shen, Yong Cheng, Zhongjun He, Wei He, Hua Wu, Maosong Sun, and Yang Liu.
  2016.
\newblock \href {https://doi.org/10.18653/v1/P16-1159} {Minimum risk training
  for neural machine translation}.
\newblock In \emph{Proceedings of the 54th Annual Meeting of the Association
  for Computational Linguistics (Volume 1: Long Papers)}, pages 1683--1692,
  Berlin, Germany. Association for Computational Linguistics.

\bibitem[{Stanojevi{\'c} and Sima{'}an(2014)}]{stanojevic-simaan-2014-fitting}
Milo{\v{s}} Stanojevi{\'c} and Khalil Sima{'}an. 2014.
\newblock \href {https://doi.org/10.3115/v1/D14-1025} {Fitting sentence level
  translation evaluation with many dense features}.
\newblock In \emph{Proceedings of the 2014 Conference on Empirical Methods in
  Natural Language Processing ({EMNLP})}, pages 202--206, Doha, Qatar.
  Association for Computational Linguistics.

\bibitem[{Tromble et~al.(2008)Tromble, Kumar, Och, and
  Macherey}]{tromble-etal-2008-lattice}
Roy Tromble, Shankar Kumar, Franz Och, and Wolfgang Macherey. 2008.
\newblock \href {https://aclanthology.org/D08-1065} {Lattice {M}inimum
  {B}ayes-{R}isk decoding for statistical machine translation}.
\newblock In \emph{Proceedings of the 2008 Conference on Empirical Methods in
  Natural Language Processing}, pages 620--629, Honolulu, Hawaii. Association
  for Computational Linguistics.

\bibitem[{Uszkoreit and Lommel(2013)}]{uszkoreit2013multidimensional}
Hans Uszkoreit and Arle Lommel. 2013.
\newblock Multidimensional quality metrics: A new unified paradigm for human
  and machine translation quality assessment.
\newblock \emph{Localization World, London}, pages 12--14.

\bibitem[{Vaswani et~al.(2017)Vaswani, Shazeer, Parmar, Uszkoreit, Jones,
  Gomez, Kaiser, and Polosukhin}]{NIPS2017_7181}
Ashish Vaswani, Noam Shazeer, Niki Parmar, Jakob Uszkoreit, Llion Jones,
  Aidan~N Gomez, {\L}ukasz Kaiser, and Illia Polosukhin. 2017.
\newblock \href
  {http://papers.nips.cc/paper/7181-attention-is-all-you-need.pdf} {{Attention
  is All you Need}}.
\newblock In I.~Guyon, U.~V. Luxburg, S.~Bengio, H.~Wallach, R.~Fergus,
  S.~Vishwanathan, and R.~Garnett, editors, \emph{Advances in Neural
  Information Processing Systems 30}, pages 5998--6008. Curran Associates, Inc.

\bibitem[{Wang et~al.(2021)Wang, Xu, Guzm{\'a}n, El-Kishky, Rubinstein, and
  Cohn}]{wang-etal-2021-easy}
Jun Wang, Chang Xu, Francisco Guzm{\'a}n, Ahmed El-Kishky, Benjamin Rubinstein,
  and Trevor Cohn. 2021.
\newblock \href {https://doi.org/10.18653/v1/2021.findings-acl.415} {As easy as
  1, 2, 3: Behavioural testing of {NMT} systems for numerical translation}.
\newblock In \emph{Findings of the Association for Computational Linguistics:
  ACL-IJCNLP 2021}, pages 4711--4717, Online. Association for Computational
  Linguistics.

\bibitem[{Zhao et~al.(2020)Zhao, Cohen, and Webber}]{zhao-etal-2020-reducing}
Zheng Zhao, Shay~B. Cohen, and Bonnie Webber. 2020.
\newblock \href {https://doi.org/10.18653/v1/2020.findings-emnlp.203} {Reducing
  quantity hallucinations in abstractive summarization}.
\newblock In \emph{Findings of the Association for Computational Linguistics:
  EMNLP 2020}, pages 2237--2249, Online. Association for Computational
  Linguistics.

\end{thebibliography}
\bibliographystyle{acl_natbib}

\clearpage

\appendix

\section{Hyperparameters for COMET Retraining}
\label{app:retrain}

We list all hyperparameters used for training the \texttt{retrain-comet-da} models with different penalties in Tables \ref{tab:train_2020}. Each model was trained on 1 NVIDIA Tesla V100 GPU.

\vspace{1cm}

\begin{table}[ht]
    \centering
    \small
    \begin{tabular}{lc}
        Hyperparameter & Value \\
        \cmidrule(lr){1-1} \cmidrule(lr){2-2} \addlinespace
        nr\_frozen\_epochs &  1  \\
        keep\_embeddings\_frozen & True \\
        optimizer & Adam \\
        encoder\_learning\_rate & 1.0e-05\\
        learning\_rate & 3.0e-05 \\
        layerwise\_decay & 0.95 \\
        encoder\_model & XLM-RoBERTa \\
        pretrained\_model & xlm-roberta-large \\
        pool & avg \\
        layer & mix \\
        dropout & 0.1 \\
        batch\_size & 2 \\
        accumulate\_grad\_batches & 8 \\
        hidden\_sizes & 3072, 1536 \\
        load\_weights\_from\_checkpoint & null \\
        min\_epochs & 2 \\
        max\_epochs & 2 \\
    \end{tabular}
    \caption{Hyperparameters used to retrain \texttt{wmt20-comet-da}.}
    \label{tab:train_2020}
\end{table}

\begin{table*}[htpb]
    \centering
    \begin{tabular}{ccccccccccc}
        &  \multicolumn{4}{c}{Numbers} & & \multicolumn{4}{c}{Named Entities}\\ 
        \cmidrule(lr){2-5} \cmidrule(lr){7-10} \addlinespace
         &  \multicolumn{2}{c}{\texttt{de-en}} &  \multicolumn{2}{c}{\texttt{en-de}} & & \multicolumn{2}{c}{\texttt{de-en}}  & \multicolumn{2}{c}{\texttt{en-de}} \\ 
        \cmidrule(lr){2-3}  \cmidrule(lr){4-5}  \cmidrule(lr){7-8}   \cmidrule(lr){9-10} \addlinespace
        reference &  93.24  &   & 93.46 & & &  n/a & & n/a &\\\addlinespace 
        alternative &  94.83 & +\phantom{0}1.59 &  95.66 & +\phantom{0}2.20 & & 73.73 &  & 77.66 &\\\addlinespace 
        beam search & 95.91 & +\phantom{0}2.67 & 95.73 & +\phantom{0}2.27 & & 71.55 & -\phantom{0}2.18 & 70.03 & -\phantom{0}7.63\\\addlinespace 
        Oracle chrF++ &  91.91 & -\phantom{0}1.33 &  93.64 & +\phantom{0}0.18 & &69.54 & -\phantom{0}4.19 & 63.59 & -14.07\\\addlinespace 
        Oracle bleu &  90.77 &  -\phantom{0}2.47 &  92.05 & -\phantom{0}1.41 & &65.73 & -\phantom{0}8.00 & 60.16 & -17.50\\\addlinespace 
        Oracle \small \texttt{wmt20-comet-da} &  90.83 & \textbf{-\phantom{0}2.41} &  88.79 & \textbf{-\phantom{0}4.67} & & 65.64 & \textbf{-\phantom{0}8.09} & 56.41 & \textbf{-21.25}\\\addlinespace 
        Oracle \small \texttt{wmt21-comet-mqm} &  91.35 & \textbf{-\phantom{0}1.89} &  86.01 & \textbf{-\phantom{0}7.45} & & 64.75 & \textbf{-\phantom{0}8.98} & 55.98 & \textbf{-21.68}\\\addlinespace 
    \end{tabular}
    \caption{Results of the automatic evaluation. ``Oracle'' means choosing the sample closest to the two reference translations. F1-scores (\%) for numbers and named entities and F1-score changes compared to the reference for numbers and alternative translation for named entities.}
    \label{tab:oracle}
\end{table*}

\begin{table*}[ht]
    \centering
    \small
    \begin{tabular}{lrccccccrcc}
         &&  \multicolumn{3}{c}{Samples as Support} & \multicolumn{3}{c}{References as Support} & & \multicolumn{2}{c}{Controls}\\ \cmidrule(lr){3-5} \cmidrule(lr){6-8} \cmidrule(lr){10-11} 
         && Numbers & NEs & Nouns &  Numbers & NEs & Nouns & & Samples & Ref.\\           \cmidrule(lr){3-3}  \cmidrule(lr){4-4}   \cmidrule(lr){5-5} \cmidrule(lr){6-6}  \cmidrule(lr){7-7}   \cmidrule(lr){8-8}   \cmidrule(lr){10-10} \cmidrule(lr){11-11} 
          \\
         \multirow{4}{*}{\textbf{de-en}} & add & -1.80 & -1.80 & \colorbox[HTML]{B2EAB1}{-1.20}  & -4.92 & -5.62 & \colorbox[HTML]{B2EAB1}{-4.41} & altern. & \phantom{-}1.11 \\
         &del & -1.70 & -1.79 & \colorbox[HTML]{B2EAB1}{-1.20} & -4.84 & -5.62 & \colorbox[HTML]{B2EAB1}{-4.41} & copy & -5.87 & -21.43\\
         &sub & -1.78 & -1.84 & \colorbox[HTML]{B2EAB1}{-1.19}  & -5.10 & -5.78 & \colorbox[HTML]{B2EAB1}{-4.44} & hallucin. & -6.71 &  -22.75\\
         &whole & -1.80 & -2.28 & \colorbox[HTML]{B2EAB1}{-1.25} & -4.92 & -6.64 & \colorbox[HTML]{B2EAB1}{-4.46} & \\\addlinespace
         \multirow{4}{*}{\textbf{en-de}} & add & -1.62 & -1.41 &  \colorbox[HTML]{B2EAB1}{-0.88}  & -4.10 & -3.56 & \colorbox[HTML]{B2EAB1}{-2.73} & altern. & -0.33 \\
         &del & -1.65 & -1.37 & \colorbox[HTML]{B2EAB1}{-0.88} & -4.24 & -3.58 & \colorbox[HTML]{B2EAB1}{-2.73} & copy & -6.02 & -20.06\\
         &sub & -1.57 & -1.41 & \colorbox[HTML]{B2EAB1}{-0.86}  &  -4.09 & -3.71 &  \colorbox[HTML]{B2EAB1}{-2.75} & hallucin. & -6.71 & -21.14\\
         &whole & -1.62 & -1.72 & \colorbox[HTML]{B2EAB1}{-0.90} & -4.10 &  -4.41 & \colorbox[HTML]{B2EAB1}{-2.79} &\\ \addlinespace
         \textbf{average} & & \textbf{-1.69} & \textbf{-1.70} & \colorbox[HTML]{B2EAB1}{\textbf{-1.05}} & \textbf{-4.54} & \textbf{-4.87} & \colorbox[HTML]{B2EAB1}{\textbf{-3.59}}
         
    \end{tabular}
    \caption{Effects of randomly adding, substituting or deleting a digit in a number or a letter in a noun or named entity (NE). Average difference to MBR score for reference (left) and 1-best beam search output (right) when using \textbf{BLEU} as the utility function. Green means both numbers and named entities have higher sensitivity than random nouns.}
    \label{tab:sensitivity_bleu}
\end{table*}

\begin{table*}[!htpb]
    \centering
    \small
    \begin{tabular}{lrccccccrcc}
         &&  \multicolumn{3}{c}{Samples as Support} & \multicolumn{3}{c}{References as Support} & & \multicolumn{2}{c}{Controls}\\ \cmidrule(lr){3-5} \cmidrule(lr){6-8} \cmidrule(lr){10-11} 
         && Numbers & NEs & Nouns &  Numbers & NEs & Nouns & & Samples & Ref.\\           \cmidrule(lr){3-3}  \cmidrule(lr){4-4}   \cmidrule(lr){5-5} \cmidrule(lr){6-6}  \cmidrule(lr){7-7}   \cmidrule(lr){8-8}   \cmidrule(lr){10-10} \cmidrule(lr){11-11} 
          \\
         \multirow{4}{*}{\textbf{de-en}} & add & -1.18 & -1.66 & \colorbox[HTML]{B2EAB1}{-1.20}  & -2.18 & -2.91 & \colorbox[HTML]{b8dfff}{-2.55} & altern. & \phantom{-}0.32\\
         &del  & -1.52 & -1.99 & \colorbox[HTML]{B2EAB1}{-1.41}  & -2.53 & -3.30 & \colorbox[HTML]{b8dfff}{-2.94} &copy& -17.18 & -32.94\\
         &sub & -1.54 & -2.00 & \colorbox[HTML]{B2EAB1}{-1.47}  & -2.74 & -3.53 & \colorbox[HTML]{b8dfff}{-3.07} & hallucin. & -22.82 & -43.39\\
         &whole & -1.91 & -4.85 & \colorbox[HTML]{b8dfff}{-2.50} & -3.25 & -8.57 & \colorbox[HTML]{b8dfff}{-5.27} & \\\addlinespace
         \multirow{4}{*}{\textbf{en-de}} & add & -0.88 & -1.25 &  \colorbox[HTML]{B2EAB1}{-0.80}  & -2.28 & -2.04 & \colorbox[HTML]{B2EAB1}{-1.52} &altern.& -0.73 &\\
         &del  & -1.10 & -1.47 & \colorbox[HTML]{B2EAB1}{-0.94}  & -1.89 & -2.37 & \colorbox[HTML]{B2EAB1}{-1.78} & copy & -19.13 & -32.68\\
         &sub  & -1.08 & -1.51 & \colorbox[HTML]{B2EAB1}{-0.96}  &  -1.87 & -2.44 &  \colorbox[HTML]{B2EAB1}{-1.81} & hallucin. & -24.96 & -42.11\\
         &whole & -1.33 & -3.72 & \colorbox[HTML]{b8dfff}{-1.98} & -2.28 &  -5.81 & \colorbox[HTML]{b8dfff}{-3.68} &  \\ \addlinespace
         \textbf{average} & & \textbf{-1.32} & \textbf{-2.31} & \colorbox[HTML]{b8dfff}{\textbf{-1.41}} & \textbf{-2.38} & \textbf{-3.87} & \colorbox[HTML]{b8dfff}{\textbf{-2.83}}
         
    \end{tabular}
    \caption{Effects of randomly adding, substituting or deleting a digit in a number or a letter in a noun or named entity (NE). Average difference to MBR score for reference (left) and 1-best beam search output (right) when using \textbf{chrf++} as the utility function. chrf++ scores are mapped to 0-100 scale for better comparison to BLEU. Green means both numbers and named entities have higher sensitivity than random nouns, blue means at least one is higher than random nouns.}
    \label{tab:sensitivity_chrf++}
\end{table*}

\section{Oracle Results for Automatic Analysis}
\label{app:oracle}
In MBR, we use machine translation metrics in an unintended way since we compare translation hypotheses against other hypotheses rather than a reference translation. To check if the results for the COMET models in our automatic analysis stem from this train-test mismatch, we also run an oracle experiment. Rather than comparing all samples against each other with MBR, we choose the sample that is most similar to the human reference translations. The results can be seen in Table \ref{tab:oracle}. Most error rates are better in the oracle setup compared to the MBR setup. Especially, the error rates for the COMET models are now closer to the non-neural metrics. However, the gap to chrF++ is still rather large, especially for named entities.

\section{MBR-based Sensitivity Analysis for BLEU and chrF++}
\label{app:sensitivity_bleu}

The MBR-based sensitivity analysis can also be used to compare COMET to non-neural metrics. The results when using BLEU or chrF++ as the utility function can be seen in Table \ref{tab:sensitivity_bleu} and Table \ref{tab:sensitivity_chrf++} respectively. We can see that with BLEU the changes made to random nouns result in smaller MBR differences than changes to numbers or named entities. For chrf++, the changes to random nouns result in smaller MBR differences than changes to named entities but slightly larger differences than changes to numbers. The cause for this may be that numbers are often shorter than named entities or nouns and a change will affect fewer n-grams. For random nouns, there may be many possible alternative translations in the samples and the references. If the random noun does not occur in the sentence we compare to, making a change to it will not affect the BLEU score and only partially the chrF++ score which can explain these results.

\begin{table*}[h]
    \centering
    \small
    \begin{tabular}{lrccccccrcc}
         &&  \multicolumn{3}{c}{Samples as Support} & \multicolumn{3}{c}{References as Support} & & \multicolumn{2}{c}{Controls}\\ \cmidrule(lr){3-5} \cmidrule(lr){6-8} \cmidrule(lr){10-11} 
         && Numbers & NEs & Nouns &  Numbers & NEs & Nouns & & Samples & Ref.\\           \cmidrule(lr){3-3}  \cmidrule(lr){4-4}   \cmidrule(lr){5-5} \cmidrule(lr){6-6}  \cmidrule(lr){7-7}   \cmidrule(lr){8-8}   \cmidrule(lr){10-10} \cmidrule(lr){11-11} 
          \\
         \multirow{4}{*}{\textbf{de-en}} & add & -0.059 & -0.067 & \colorbox[HTML]{F9CBD0}{-0.230} & -0.116 & -0.135 & \colorbox[HTML]{F9CBD0}{-0.386} & altern. & \phantom{-}0.021 & \\
         &del & -0.048 & -0.053 & \colorbox[HTML]{F9CBD0}{-0.199}  & -0.092 & -0.105 & \colorbox[HTML]{F9CBD0}{-0.326} & copy & -0.778 & -0.690\\
         &sub  & -0.028 & -0.065 & \colorbox[HTML]{F9CBD0}{-0.242} & -0.054 & -0.146 & \colorbox[HTML]{F9CBD0}{-0.403}& hallucin. & -1.081  & -1.720\\
         &whole & -0.082 & -0.127 & \colorbox[HTML]{F9CBD0}{-0.287} & -0.151 & -0.250 & \colorbox[HTML]{F9CBD0}{-0.493} & \\\addlinespace
         \multirow{4}{*}{\textbf{en-de}} & add & -0.040 & -0.044 &  \colorbox[HTML]{F9CBD0}{-0.153} & -0.083 & -0.107 & \colorbox[HTML]{F9CBD0}{-0.260} & altern. & -0.015 \\
         &del  & -0.046 & -0.038 & \colorbox[HTML]{F9CBD0}{-0.117}  & -0.080 & -0.083 & \colorbox[HTML]{F9CBD0}{-0.211} & copy & -1.513 & -1.625\\
         &sub  & -0.015 & -0.051 & \colorbox[HTML]{F9CBD0}{-0.169} &  -0.034 & -0.111 &  \colorbox[HTML]{F9CBD0}{-0.277} & hallucin. &-1.402 & -1.891\\
         &whole & -0.055 & -0.106 & \colorbox[HTML]{F9CBD0}{-0.353} &  -0.109 &  -0.197 & \colorbox[HTML]{F9CBD0}{-0.541} &\\ \addlinespace
         \textbf{average} & & \textbf{-0.047} & \textbf{-0.069} & \colorbox[HTML]{F9CBD0}{\textbf{-0.219}} & \textbf{-0.090} & \textbf{-0.108} & \colorbox[HTML]{F9CBD0}{\textbf{-0.362}}
         
    \end{tabular}
    \caption{Effects of randomly adding, substituting or deleting a digit in a number or a letter in a noun or named entity (NE) compared to the controls (alternative translation, copy of the source or hallucination). The numbers show the average difference to the MBR score for the reference (left) and 1-best beam search output (right) when using \texttt{retrain-comet-da} with a \textbf{penalty of -0.2} as the utility function. Red means the sensitivity for random nouns is larger than for both numbers and named entities.}
    \label{tab:sensitivity_0.2}
\end{table*}

\begin{table*}[h]
    \centering
    \small
    \begin{tabular}{lrccccccrcc}
         &&  \multicolumn{3}{c}{Samples as Support} & \multicolumn{3}{c}{References as Support} & & \multicolumn{2}{c}{Controls}\\ \cmidrule(lr){3-5} \cmidrule(lr){6-8} \cmidrule(lr){10-11} 
         && Numbers & NEs & Nouns &  Numbers & NEs & Nouns & & Samples & Ref.\\           \cmidrule(lr){3-3}  \cmidrule(lr){4-4}   \cmidrule(lr){5-5} \cmidrule(lr){6-6}  \cmidrule(lr){7-7}   \cmidrule(lr){8-8}   \cmidrule(lr){10-10} \cmidrule(lr){11-11} 
          \\
         \multirow{4}{*}{\textbf{de-en}} & add & -0.243 & -0.229 & \colorbox[HTML]{F9CBD0}{-0.337} & -0.417 & -0.382 & \colorbox[HTML]{F9CBD0}{-0.523} & altern. & \phantom{-}0.026 & \\
         &del & -0.217 & -0.180 & \colorbox[HTML]{F9CBD0}{-0.261}  & -0.380 & -0.295 & \colorbox[HTML]{F9CBD0}{-0.410} & copy & -0.471 & -0.409\\
         &sub  & -0.152 & -0.223 & \colorbox[HTML]{F9CBD0}{-0.347} & -0.256 & -0.402 & \colorbox[HTML]{F9CBD0}{-0.542}& hallucin. & -1.076  & -1.724\\
         &whole & -0.312 & -0.197 & \colorbox[HTML]{F9CBD0}{-0.320} & -0.529 & -0.374 & \colorbox[HTML]{b8dfff}{-0.521} & \\\addlinespace
         \multirow{4}{*}{\textbf{en-de}} & add & -0.224 & -0.210 &  \colorbox[HTML]{F9CBD0}{-0.231} & -0.405 & -0.379 & \colorbox[HTML]{b8dfff}{-0.379} & altern. & -0.017 \\
         &del  & -0.197 & -0.156 & \colorbox[HTML]{B2EAB1}{-0.148}  & -0.319 & -0.261 & \colorbox[HTML]{b8dfff}{-0.262} & copy & -1.142 & -1.133\\
         &sub  & -0.129 & -0.196 & \colorbox[HTML]{F9CBD0}{-0.250} &  -0.213 & -0.352 &  \colorbox[HTML]{F9CBD0}{-0.392} & hallucin. &-1.370 & -1.895\\
         &whole & -0.275 & -0.196 & \colorbox[HTML]{F9CBD0}{-0.339} &  -0.493 &  -0.351 & \colorbox[HTML]{F9CBD0}{-0.516} &\\ \addlinespace
         \textbf{average} & & \textbf{-0.219} & \textbf{-0.198} & \colorbox[HTML]{F9CBD0}{\textbf{-0.279}} & \textbf{-0.377} & \textbf{-0.350} & \colorbox[HTML]{F9CBD0}{\textbf{-0.511}}
         
    \end{tabular}
    \caption{Effects of randomly adding, substituting or deleting a digit in a number or a letter in a noun or named entity (NE) compared to the controls (alternative translation, copy of the source or hallucination). The numbers show the average difference to the MBR score for the reference (left) and 1-best beam search output (right) when using \texttt{retrain-comet-da} with a \textbf{penalty of -0.5} as the utility function. Red means the sensitivity for random nouns is larger than for both numbers and named entities, blue means at least one is higher than random nouns and green means both numbers and named entities have higher sensitivity than random nouns.}
    \label{tab:sensitivity_0.5}
\end{table*}

\begin{table*}[h]
    \centering
    \small
    \begin{tabular}{lrccccccrcc}
         &&  \multicolumn{3}{c}{Samples as Support} & \multicolumn{3}{c}{References as Support} & & \multicolumn{2}{c}{Controls}\\ \cmidrule(lr){3-5} \cmidrule(lr){6-8} \cmidrule(lr){10-11} 
         && Numbers & NEs & Nouns &  Numbers & NEs & Nouns & & Samples & Ref.\\           \cmidrule(lr){3-3}  \cmidrule(lr){4-4}   \cmidrule(lr){5-5} \cmidrule(lr){6-6}  \cmidrule(lr){7-7}   \cmidrule(lr){8-8}   \cmidrule(lr){10-10} \cmidrule(lr){11-11} 
          \\
         \multirow{4}{*}{\textbf{de-en}} & add & -0.435 & -0.412 & \colorbox[HTML]{B2EAB1}{-0.401} & -0.706 & -0.687 & \colorbox[HTML]{B2EAB1}{-0.617} & altern. & \phantom{-}0.024 & \\
         &del & -0.385 & -0.331 & \colorbox[HTML]{B2EAB1}{-0.293}  & -0.655 & -0.526 & \colorbox[HTML]{B2EAB1}{-0.450} & copy & -0.306 & -0.234\\
         &sub  & -0.305 & -0.547 & \colorbox[HTML]{b8dfff}{-0.394} & -0.472 & -0.667 & \colorbox[HTML]{b8dfff}{-0.614}& hallucin. & -1.225  & -1.962\\
         &whole & -0.547 & -0.267 & \colorbox[HTML]{b8dfff}{-0.320} & -0.889 & -0.495 & \colorbox[HTML]{b8dfff}{-0.539} & \\\addlinespace
         \multirow{4}{*}{\textbf{en-de}} & add & -0.381 & -0.337 &  \colorbox[HTML]{b8dfff}{-0.337} & -0.657 & -0.635 & \colorbox[HTML]{B2EAB1}{-0.575} & altern. & -0.015 \\
         &del  & -0.355 & -0.254 & \colorbox[HTML]{B2EAB1}{-0.230}  & -0.614 & -0.457 & \colorbox[HTML]{B2EAB1}{-0.402} & copy & -0.852 & -0.755\\
         &sub  & -0.264 & -0.322 & \colorbox[HTML]{F9CBD0}{-0.351} &  -0.437 & -0.585 &  \colorbox[HTML]{b8dfff}{-0.570} & hallucin. &-1.498 & -2.046\\
         &whole & -0.470 & -0.271 & \colorbox[HTML]{b8dfff}{-0.370} &  -0.827 &  -0.484 & \colorbox[HTML]{b8dfff}{-0.550} &\\ \addlinespace
         \textbf{average} & & \textbf{-0.393} & \textbf{-0.343} & \colorbox[HTML]{B2EAB1}{\textbf{-0.337}} & \textbf{-0.657} & \textbf{-0.567} & \colorbox[HTML]{B2EAB1}{\textbf{-0.540}}
         
    \end{tabular}
    \caption{Effects of randomly adding, substituting or deleting a digit in a number or a letter in a noun or named entity (NE) compared to the controls (alternative translation, copy of the source or hallucination). The numbers show the average difference to the MBR score for the reference (left) and 1-best beam search output (right) when using \texttt{retrain-comet-da} with a \textbf{penalty of -0.8} as the utility function. Red means the sensitivity for random nouns is larger than for both numbers and named entities, blue means at least one is higher than random nouns and green means both numbers and named entities have higher sensitivity than random nouns.}
    \label{tab:sensitivity_0.8}
\end{table*}

\section{Retraining with Different Penalties}
\label{app:punish}

Tables \ref{tab:sensitivity_0.2}, \ref{tab:sensitivity_0.5}, \ref{tab:sensitivity_0.8} show the results of the sensitivity analysis for the retrained models with penalties of -0.2, -0.5 and -0.8 respectively. The difference between the sensitivity scores for numbers / named entities and for random nouns becomes smaller as the penalty increases. With a penalty of -0.8, we see that for most error types the sensitivity scores for random nouns are either lower than either (blue) or both (green) for numbers and named entities. Note that the differences in MBR score compared to the reference (left) and the 1-best beam search output (right) also become larger as the penalties increase. However, this does not affect on the models' ability to score real translations as we confirm in Section \ref{app:corr}.

\section{Retraining with Different Amounts of Synthetic Data}
\label{app:synthetic}

Aside from varying the penalties for retraining COMET (see Appendix \ref{app:punish}), we can also vary the amount of synthetic data. Using the best performing penalty from before (0.8), we run experiments with 0\%, 10\%, 25\%, 40\%, 55\%, 70\%, 85\% and 100\% synthetic data for retraining COMET. Note that 0\% corresponds to the original \texttt{wmt20-comet-da} model and 10\% corresponds to \texttt{retrain-comet-da} in the main paper experiments. We evaluate these models based on two factors: 1) the average difference in sensitivity between the number and named entity error types and the random nouns (corresponding to an average over the individual columns in Figure \ref{fig:diff_nouns}) and 2) the change in Pearson correlation compared to \texttt{wmt20-comet-da}. The first measure indicates how the retrained models' sensitivity to numbers and named entities changes compared to random nouns with increased synthetic data. The second measure shows whether an increased amount of synthetic data reduces the agreement with human judgements (this is computed as described in Appendix \ref{app:corr}).

Figure \ref{fig:diff_sensitivity} shows that with an increased percentage of synthetic data, the difference between sensitivity towards nouns and named entities and towards random noun changes first becomes smaller (at 10\% synthetic). When we further increase the amount of synthetic data, this improvement gradually decreases as the model sees less and less contrasting examples and more and more only examples with number mismatches. 

\begin{figure}
    \centering
    \includegraphics[width=0.48\textwidth]{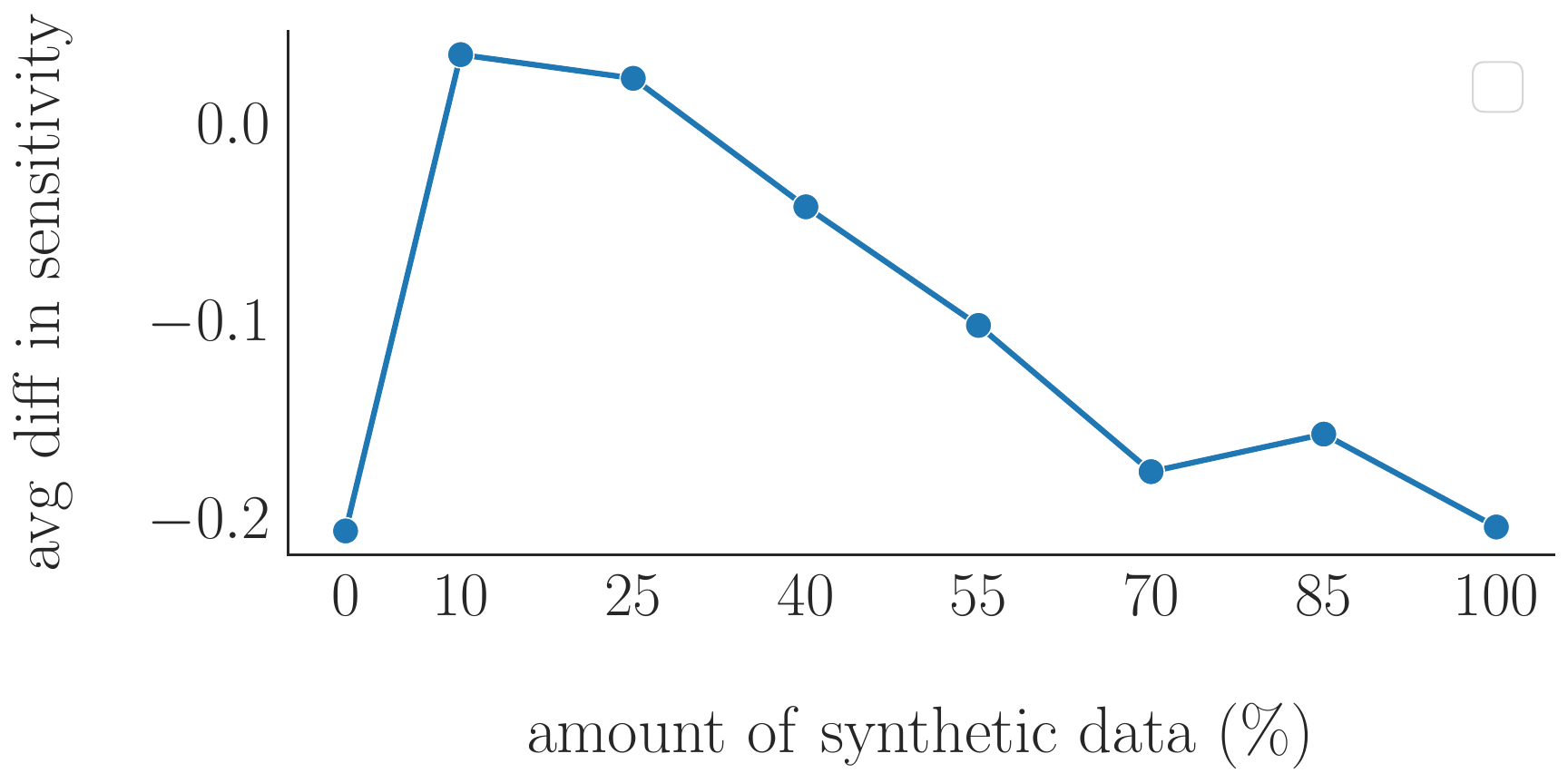}
    \caption{The average difference in sensitivity between the noun/named entity error categories and their corresponding random noun error categories. The x-axis shows how the difference changes as the amount of synthetic data is increased.}
    \label{fig:diff_sensitivity}
\end{figure}

\begin{figure}
    \centering
    \includegraphics[width=0.48\textwidth]{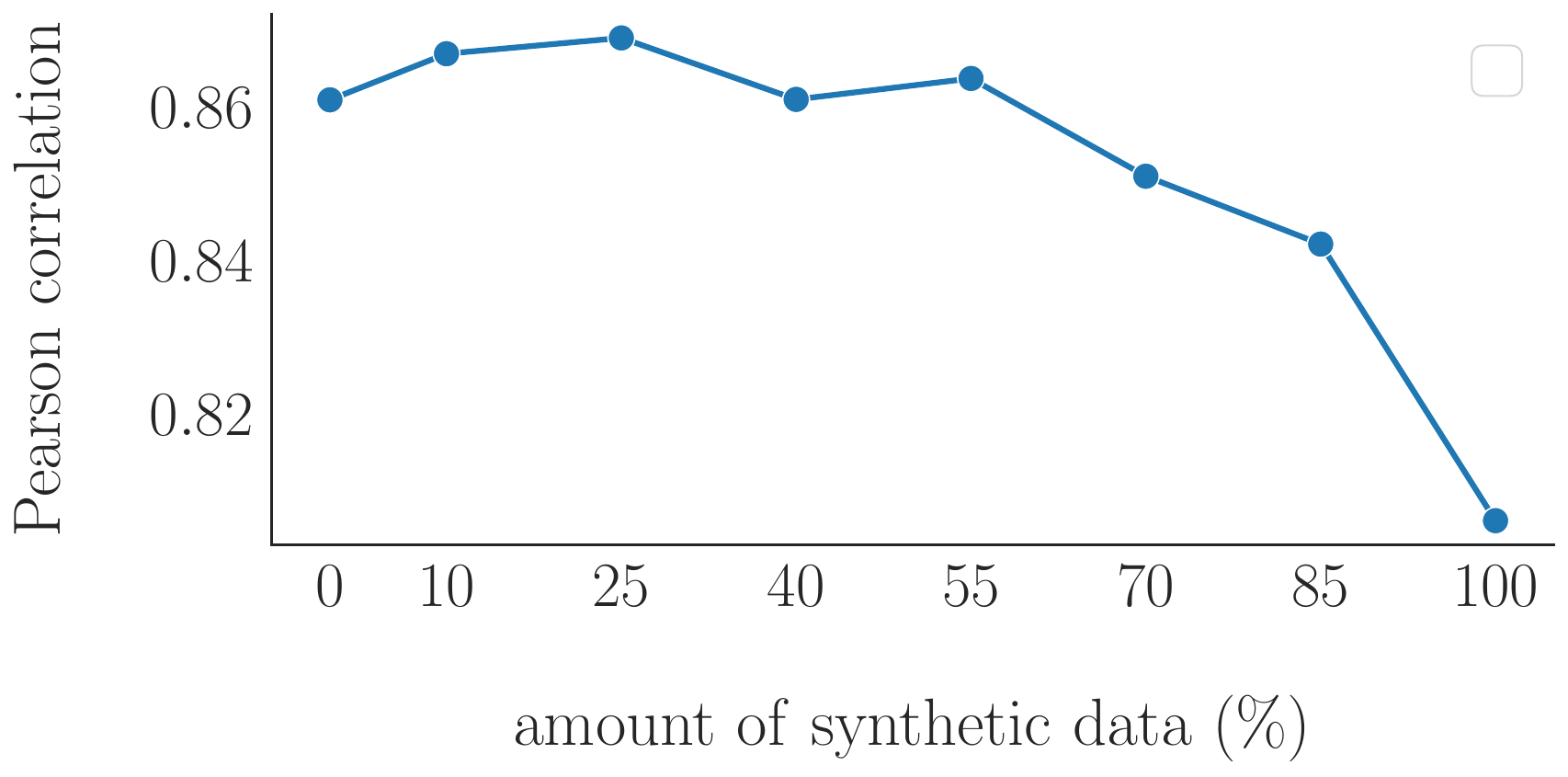}
    \caption{The correlation with human judgements evaluated as described in Appendix \ref{app:corr}. The x-axis shows how the correlation changes as the amount of synthetic data is increased.}
    \label{fig:correlation}
\end{figure}

Increasing the amount of synthetic data during retraining also has an effect on the correlation with human judgements. We show this in Figure \ref{fig:correlation}. Similarly to the difference in sensitivity, the correlation with human judgements also improves with small amounts of synthetic data (10\% and 25\%) but then decreases slowly as the amount of synthetic data is increased further. These additional experiments show that using 10\% of synthetic data is a sensible choice for our main experiments.

\section{Correlation with Human Evaluators}
\label{app:corr}

We use our retrained \texttt{retrain-comet-da} models to score all systems that are part of the WMT 2020 metrics shared task evaluation \citep{mathur-etal-2020-results}.\footnote{We run the \texttt{run\_ref\_metrics.sh} script provided at \url{https://drive.google.com/drive/folders/1n_alr6WFQZfw4dcAmyxow4V8FC67XD8p}} Then, we use the official evaluation script\footnote{\url{https://github.com/WMT-Metrics-task/wmt20-metrics}} from the WMT 2020 shared task to compute the system-level Pearson correlation for our retrained models. The results can be seen in Table \ref{tab:corr}. We also ensure that evaluation setup results in the same scores as in the WMT 2020 publication \citep{mathur-etal-2020-results} when we use \texttt{wmt20-comet-da} to score the systems. For most language pairs, all models reach an almost identical correlation with human assessments.

\begin{table*}[!ht]
    \centering
    \small
    \begin{tabular}{ccccc}
    &  \footnotesize \texttt{wmt20-comet-da} & \multicolumn{3}{c}{\footnotesize \texttt{retrain-comet-da}}\\ 
   \cmidrule(lr){2-2}  \cmidrule(lr){3-5} \addlinespace
    &  &  -0.2  & -0.5 & -0.8\\ 
     \cmidrule(lr){3-3}  \cmidrule(lr){4-4} \cmidrule(lr){5-5}\addlinespace
        en-cs & 0.978 & 0.981 & 0.981 & 0.981\\ \addlinespace
        en-de & 0.972 & 0.971 & 0.965 & 0.963\\ \addlinespace
        en-ja & 0.974 & 0.987 & 0.974 & 0.982\\ \addlinespace
        en-pl & 0.981 & 0.983 & 0.985 & 0.983\\ \addlinespace
        en-ru & 0.925 & 0.863 & 0.900 & 0.918\\ \addlinespace
        en-ta & 0.944 & 0.948 & 0.949 & 0.954\\ \addlinespace
        en-zh & 0.007 & 0.026 & 0.034 & 0.049\\ \addlinespace
        en-iu & 0.860 & 0.861 & 0.851 & 0.873\\ \addlinespace \addlinespace
        cs-en & 0.783 & 0.799 & 0.798 & 0.808\\ \addlinespace
        de-en & 0.998 & 0.996 & 0.995 & 0.997\\ \addlinespace
        ja-en & 0.964 & 0.966 & 0.968 & 0.968\\ \addlinespace
        pl-en & 0.591 & 0.570 & 0.570 & 0.563\\ \addlinespace
        ru-en & 0.923 & 0.924 & 0.921 & 0.925\\ \addlinespace
        ta-en & 0.880 & 0.888 & 0.887 & 0.890\\ \addlinespace
        zh-en & 0.952 & 0.952 & 0.942 & 0.951\\ \addlinespace
        iu-en & 0.852 & 0.878 & 0.866 & 0.880\\ \addlinespace
        km-en & 0.971 & 0.981 & 0.981 & 0.974\\ \addlinespace
        ps-en & 0.941 & 0.951 & 0.949 & 0.945\\ \addlinespace \hline \addlinespace
        avg diff & & +0.0016 & -0.0006 & +0.0060
    \end{tabular}
    \caption{Pearson correlation of to-and-from-English system-level COMET scores with DA human assessments. Last row shows the average difference to the original \texttt{wmt20-comet-da} model. Results with \texttt{wmt20-comet-da} corresponding to ``COMET'' in Tables 5 and 6 in \citet{mathur-etal-2020-results}.}
    \label{tab:corr}
\end{table*}

\end{document}